\documentclass[preprint, 12pt, square]{elsarticle}  

\usepackage[left=2.cm,right=2.cm,top=2.54cm,bottom=2.54cm]{geometry}

\usepackage{floatrow}
\usepackage{graphicx}
\usepackage[]{subfig} 
\usepackage{amsmath,bm}
\usepackage{color}
\usepackage{enumitem}
\usepackage{multirow}
\usepackage{amssymb}
\usepackage{epstopdf}
\usepackage{epsfig}
\usepackage[table,xcdraw]{xcolor}

\usepackage{algorithm}  
\usepackage{algorithmic}
\usepackage{eqparbox}

\usepackage{tikz}
\usetikzlibrary{positioning}

\usepackage{natbib}
\setcitestyle{authoryear,colon,round}

\usepackage{bigstrut}

\usepackage{caption}

\usepackage{hyperref}

\usepackage{xr-hyper}

\usepackage{url}

\definecolor{lightyellow}{rgb}{0.95, 0.95, 0.85}

\renewcommand{\algorithmicrequire}{\textbf{Input:}}  
\renewcommand{\algorithmicensure}{\textbf{Output:}} 

\journal{ }



\begin{document}

\begin{frontmatter}

\title{Optimization of geological carbon storage operations with multimodal latent dynamic model and deep reinforcement learning}

\author[1]{Zhongzheng Wang}
\author[2]{Yuntian Chen\corref{mycorrespondingauthor}}
\cortext[mycorrespondingauthor]{Corresponding author}
\ead{ychen@eitech.edu.cn}
\author[3]{Guodong Chen}
\author[2,4]{Dongxiao Zhang\corref{mycorrespondingauthor}}
\ead{dzhang@eitech.edu.cn}

\address[1]{BIC-ESAT, ERE, and SKLTCS, College of Engineering, Peking University, Beijing 100871, P.R. China}
\address[2]{Ningbo Institute of Digital Twin, Eastern Institute of Technology, Ningbo, Zhejiang 315200, P.R. China}
\address[3]{Department of Earth Sciences, The University of Hong Kong, Hong Kong 999077, P.R. China}
\address[4]{Department of Mathematics and Theories, Peng Cheng Laboratory, Shenzhen, Guangdong 518000, P.R. China}

\begin{abstract}
\noindent{Identifying the time-varying control schemes that maximize storage performance is critical to the commercial deployment of geological carbon storage (GCS) projects. However, the optimization process typically demands extensive resource-intensive simulation evaluations, which poses significant computational challenges and practical limitations. In this study, we presented the multimodal latent dynamic (MLD) model, a novel deep learning framework for fast flow prediction and well control optimization in GCS operations. The MLD model implicitly characterizes the forward compositional simulation process through three components: a representation module that learns compressed latent representations of the system, a transition module that approximates the evolution of the system states in the low-dimensional latent space, and a prediction module that forecasts the flow responses for given well controls. A novel model training strategy combining a regression loss and a joint-embedding consistency loss was introduced to jointly optimize the three modules, which enhances the temporal consistency of the learned representations and ensures multi-step prediction accuracy. Unlike most existing deep learning models designed for systems with specific parameters, the MLD model supports arbitrary input modalities, thereby enabling comprehensive consideration of interactions between diverse types of data, including dynamic state variables, static spatial system parameters, rock and fluid properties, as well as external well settings. Since the MLD model mirrors the structure of a Markov decision process (MDP) that computes state transitions and rewards (i.e., economic calculation for flow responses) for given states and actions, it can serve as an interactive environment to train deep reinforcement learning agents. Specifically, the soft actor-critic (SAC) algorithm was employed to learn an optimal control policy that maximizes the net present value (NPV) from the experiences gained by continuous interactions with the MLD model. The efficacy of the proposed approach was first compared against commonly used simulation-based evolutionary algorithm and surrogate-assisted evolutionary algorithm on a deterministic GCS optimization case, showing that the proposed approach achieves the highest NPV, while reducing the required computational resources by more than 60\%. The framework was further applied to the generalizable GCS optimization case. The results indicate that the trained agent is capable of harnessing the knowledge learned from previous scenarios to provide improved decisions for newly encountered scenarios, demonstrating promising generalization performance.}
\end{abstract}


\end{frontmatter}

\section{Introduction}
\label{sec1}
Geological carbon storage (GCS) is considered as one of the most promising technologies for reducing greenhouse gas emissions and mitigating climate change \citep{zahasky2020global, middleton2020identifying}. In GCS, the carbon dioxide (CO$_2$) captured from anthropogenic sources (e.g., power generation or industrial facilities) is compressed into the super-critical fluid status and then injected into underground geological formations (e.g., saline aquifers or depleted hydrocarbon reservoirs) for long-term sequestration \citep{celia2015status, tang2021deep}. In recent years, the implementation of carbon tax incentives has attracted interested parties to invest in GCS projects \citep{sun2020optimal}. However, the identification of safe and cost-effective operational strategies is confronted with significant challenges due to the highly complex physical and chemical processes. During GCS operations, different well control schemes lead to large variations in CO$_2$ plume evolution and migration. If not designed properly, it may reduce dynamic storage capacity and result in potential environmental impacts \citep{cihan2015optimal}. Consequently, for the actual deployment of GCS projects, it is imperative to optimize time-varying well controls to improve overall storage performance.

High-fidelity numerical simulations are widely applied to describe subsurface flow behaviors and establish correlations between well control schemes and storage performance metrics (objective functions) \citep{zhang2013numerical, kawata2017some}. In recent years, domain practitioners have proposed various simulation-based optimization methods, which can be broadly categorized into gradient-based algorithms and derivative-free algorithms. Among them, gradient-based algorithms demonstrate high computational efficiency since they directly utilize gradient information to determine the optimal direction of moving a candidate solution \citep{zandvliet2008adjoint, Fonseca2017ASS}. However, the gradient information required for optimization is often difficult or impossible to acquire from the packaged simulator \citep{zhong2022historical}. As an alternative, derivative-free algorithms, such as differential evolution and particle swarm optimization, overcome this obstacle well as they proceed using stochastic evolutionary operators without calculating the derivatives \citep{foroud2018comparative, musayev2023optimization}. Such methods enjoy the flexibility of harnessing existing simulation software and have been extensively used for case studies \citep{lu2022bayesian, zou2023integrated}. Nevertheless, this kind of algorithm typically suffers from relatively slow convergence, especially for cases with high-dimensional search spaces \citep{wang2023surrogate}. In addition to their own strengths and weaknesses, the aforementioned algorithms have a common fundamental issue: they independently solve each task, and the optimization of a new task must be performed from scratch even for a related one. 

The rapid development of deep reinforcement learning (DRL) has introduced a promising direction, which recent studies refer to as generalizable optimization \citep{he2022deep, wang2022deep}. Instead of searching for a single solution, DRL aims to learn a control policy through the interaction of an agent with its environment \citep{arulkumaran2017deep, zhang2022training}. The policy is a function that maps perceived states of the environment to actions to be taken, which can be optimized by solving a Markov decision process (MDP). With the optimal policy, the DRL agent can provide improved solutions to new optimization tasks without resolving the problem again \citep{cao2020model}. Currently, DRL has been increasingly applied to deal with a variety of challenging sequential decision-making problems in subsurface energy system design, such as optimization of hydrocarbon production \citep{miftakhov2020deep, he2022deep, dixit2022stochastic, wang2023hierarchical}, estimation of uncertain parameters \citep{li2021reinforcement}, and closed-loop control optimization \citep{nasir2023deep, nasir2023practical}. However, there are not many applications of DRL in CO$_2$ storage management. \cite{sun2020optimal} employed the deep Q-learning algorithm to identify the optimal CO$_2$ injection profiles for both a passive management scenario and an active management scenario. Despite promising results, only a two-dimensional (2D) case was considered and the experiments were limited to a discrete action space involving a single type of decision variable. One of the primary reasons limiting the widespread adoption of DRL in this field is the huge computational challenges. The DRL agent necessitates a massive number of simulation evaluations to learn the control policies. Unfortunately, due to the heterogeneity of porous media and the coupled-physics nature, a single simulation run may take minutes or even hours as the scale of the numerical model increases. As a consequence, how to speed up forward evaluation while maintaining satisfactory prediction accuracy is the key to scaling DRL to practical applications. 

In the general context of the fluid flow and transport in porous media, deep learning-based surrogate models have recently emerged as attractive tools to complement or replace numerical simulators \citep{wang2020deep, zhong2021deep, tang2022deep}. They are capable of capturing the high-dimensional spatial–temporal information associated with dynamic physical systems with relatively limited training data \citep{zhu2018bayesian, tang2020deep}. To treat optimization problems, numerous efforts have been undertaken to construct efficient and accurate surrogates for subsurface flow simulation with time-varying well controls. One line of research seeks to transform the forward flow prediction problem into different types of regression tasks, such as sequence-to-sequence regression \citep{kim2021recurrent} and image-to-sequence \citep{kim2023convolutional}. Such methods can predict the system states at a series of time steps for given inputs. However, they cannot predict the state transitions of the system and are thus not applicable to sequential optimization problems. One approach well suited to address this challenge is autoregressive methods \citep{mo2019deep, sun2020optimal}. The deep autoregressive models can learn and approximate state transitions between consecutive time steps. However, they typically need to update the state of each discretized cell at each time step, which makes the training and inference speed relatively slow \citep{wu2022learning}. To immediate this issue, recent works developed the deep learning-based reduced-order modeling (ROM) frameworks \citep{jin2020deep, huang2024application}, in which the autoregressive strategy was applied to a low-dimensional latent space. ROM procedures rely on reconstructing the system states to shape latent representations and to provide supervised signals for model training. Results on oil-water reservoir simulation problems validated the effectiveness and efficiency of the proposed approaches. However, they are restricted to providing reliable predictions for a deterministic scenario. When the numerical model parameters or fluid properties change, the model has to be built from scratch, which limits its further applications.

Considering the above issues, a salient question naturally arises: are there more general deep learning models that can comprehensively consider the interaction between various types of data (including dynamic spatial state variables, static spatial system parameters, static vector-type fluid properties, and time-varying well controls), and can accurately approximate the evolution of the system states over long time steps? 

In this study, we answered this question affirmatively. In particular, we developed the multimodal latent dynamic (MLD) model, a novel deep learning framework that implicitly describes the forward numerical simulation process. Specifically, the MLD model consists of three components: (1) a representation module that compresses multimodal input features into low-dimensional latent representations, (2) a transition module that evolves the system states in the latent space, and (3) a prediction module that forecasts flow responses for given well specifications. To ensure multi-step prediction accuracy, we further introduced a novel training strategy combining a regression loss with a joint-embedding consistency loss to jointly optimize the three modules. The key insights behind the MLD model include three aspects. First, the distinct types of data involved in numerical simulations can be viewed as multimodal features, which can be further abstracted into higher-level representations. Second, approximating the evolution of the system states in the low-dimensional latent space is much more efficient and easier than in the original high-dimensional space. Indeed, the idea of "latent evolution" has been adopted in many other domains, such as robotics \citep{hafner2019learning, hansen2022temporal}, fluid dynamics \citep{wu2022learning}, and computer vision \citep{watters2017visual, udrescu2021symbolic}. Third, the MLD model learns latent representations of the system purely from the flow responses without reconstructing the high-dimensional system states, which mitigates error accumulation while enabling it to support arbitrary input modalities. After validating the MLD model against simulation results, it was incorporated into the DRL framework, in which a soft actor-critic (SAC) \citep{haarnoja2018soft} agent was trained to learn the most rewarding control policy from the experiences obtained by continuous interactions with the MLD model. The resulting algorithm, denoted as MSDRL, represents the first-of-its-kind framework to combine the multimodal deep learning models with DRL for GCS optimization. The efficacy of MSDRL was successfully demonstrated using both deterministic optimization scenario and generalizable optimization scenario.

The rest of the paper proceeds as follows. Section ~\ref{sec2} provides the general forward compositional simulation process and describes how to formulate the GCS optimization problem in the DRL framework. Section ~\ref{sec3} introduces the construction of the MLD model and the SAC agent model as well as the overall optimization workflow. Subsequently, the performance evaluation of the MLD model and the optimization results for 3D GCS cases are presented in Section ~\ref{sec4}. Finally, section ~\ref{sec5} discusses and summarizes this work, and provides insights into future research directions.

\section{Optimization modeling of GCS process}
\label{sec2}

\subsection{Forward compositional simulation}
This work focuses on the case of CO$_2$ storage in saline aquifers as it has the largest identified storage potential and offers the main solution to scale GCS deployment \citep{celia2015status}. Specifically, the supercritical CO$_2$ is injected into the deep saline aquifers and the brine is extracted from the production wells. Compositional simulations are widely used to model forward flow processes. For the considered multiphase flow system, the mass conservation equation for each component $i$ in present phase $\alpha$ can be given as:
\begin{equation} 
    \label{mass}
    \frac{\partial }{\partial t}\left( \sum\limits_{\alpha }{\phi {{S}_{\alpha }}{{\rho }_{\alpha }}X_{\alpha }^{i}} \right)+\nabla \cdot \sum\limits_{\alpha }{\left( {{\rho }_{\alpha }}X_{\alpha }^{i}{{\mathbf{v}}_{\alpha }}-\phi {{S}_{\alpha }}D_{\alpha }^{i}\mathbf{I}\nabla \left( {{\rho }_{\alpha }}X_{\alpha }^{i} \right) \right)}={{q}^{i}}
\end{equation}
where $t$ is the time. $\alpha$ is the fluid phase. $\mathsf{\phi}$ is the rock porosity. ${{\rho}_{\alpha}}$ is the phase density. ${{S}_{\alpha}}$ is the phase saturation, which is constrained by $\sum\nolimits_{\alpha}{{{S}_{\alpha}}}=1$. $X_{\alpha }^{i}$ and $D_{\alpha }^{i}$ are the mass fraction and normal diffusion coefficient of component $i$ in phase $\alpha$, respectively. ${{q}^{i}}$ is the source/sink term of component $i$. ${{\mathbf{v}}_{\alpha}}$ is the Darcy velocity, which is expressed as:
\begin{equation} 
    {{\mathbf{v}}_{\alpha}}=-\frac{\mathbf{k}{{k}_{r\alpha }}}{{{\mu }_{\alpha }}}\left( \nabla {{P}_{\alpha }}-{{\rho }_{\alpha }}\text{g}\nabla z \right)
    \label{darcy}
\end{equation}
where $\mathbf{k}$ is the absolute permeability. ${{k}_{r\alpha}}$ is the phase relative permeability. ${{\mu }_{\alpha}}$ is the phase viscosity. ${{P}_{\alpha}}$ is the phase pressure. $\text{g}$ is the gravitational acceleration, and $z$ is the depth.

The evolution of system states can be described by discretizing and solving the above governing equations, given initial and boundary conditions. Without loss of generality, the forward simulation can be described as a dynamic model, which contains a state equation Eq.~\ref{stateeq} and an output equation Eq.~\ref{outputeq}:
\begin{equation} 
    {{\mathbf{u}}_{t+\text{1}}}=g\left( {{\mathbf{u}}_{t}},\mathbf{m},\mathbf{o},{{\mathbf{a}}_{t}} \right)
    \label{stateeq}
\end{equation}
\begin{equation} 
    {{\mathbf{d}}_{t}}=h\left({{\mathbf{u}}_{t}},\mathbf{m},\mathbf{o},{{\mathbf{a}}_{t}} \right)
    \label{outputeq}
\end{equation}
where ${\mathbf{u}}_{t}$ denotes the dynamic spatial state variables (e.g., phase saturation and pressure) at the time $t$. $\mathbf{m}$ denotes the static spatial system parameters (e.g., permeability and porosity). $\mathbf{o}$ denotes the static vector-type parameters, such as relative permeability and other fluid properties. ${\mathbf{a}}_{t}$ represents the well controls at the time $t$. ${\mathbf{d}}_{t}$ represents the flow responses (e.g., brine production rate and CO$_2$ injection rate) at the time $t$. $g$ and $h$ are the nonlinear functions, describing the numerical simulation process.

\subsection{Optimization problem formulation}
\label{problem}
In the context of the implementation of the carbon tax policy, the feasibility of the GCS project deployment can be assessed by the economic benefits. Referring to previous studies \citep{kawata2017some, sun2021optimization}, we chose the net present value (NPV) as the storage performance metric to evaluate the quality of different operation strategies. For GCS processes, the NPV can be calculated by Eq.~\ref{npv}, in which carbon tax credits and brine treatment costs are taken into account. The objective of optimization is to determine the time-varying well controls that maximize NPV over the lifespan of the project.
\begin{equation} 
    {\rm NPV}=\sum\limits_{n=1}^{H}{\frac{{{r}_{c}}\cdot {{Q}_{c,n}}-{{r}_{b}}\cdot {{Q}_{b,n}}}{{{\left( 1+b \right)}^{{{t}_{n}}}}}\Delta {{t}_{n}}}
    \label{npv}
\end{equation}
where ${Q}_{c,n}$ and ${Q}_{b,n}$ are the average CO$_2$ injection rate [$m^{3}$/day] and brine production rate [$m^{3}$/day] during the $n$th time step, respectively. $r_{c}$ and $r_{b}$ are the profit from CO$_2$ injected [USD/$m^{3}$] and brine treatment cost [USD/$m^{3}$], respectively. $\Delta t_n$ is the length of the $n$th time step [day]. $b$ is the annual discount rate. $t_n$ is the time at the end of the $n$th time step [year]. $H$ is the total number of time steps.

\begin{figure}[!htb]
    \centering
    \includegraphics[width=0.45\textwidth]{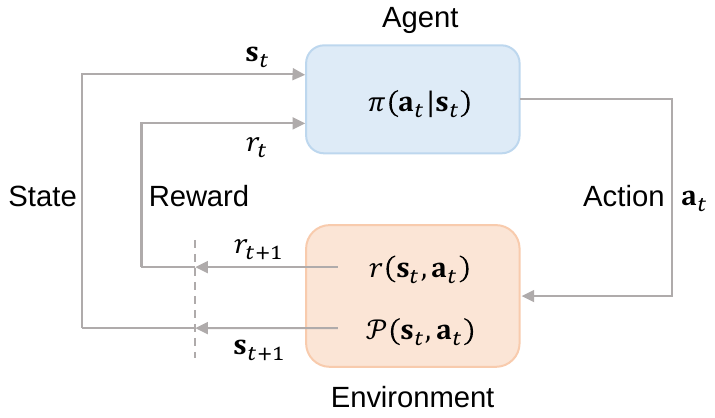}
    \caption{Schematic of agent–environment interaction in an MDP.}
    \label{/MDP}
\end{figure}

Considering the sequential decision-making nature of GCS operations, we adopted DRL to solve the well control optimization problem. A commonly used approach in DRL is to formulate the target task as a Markov decision process (MDP), which provides a fundamental framework for the agent-environment interaction \citep{sutton2018reinforcement}. In general, an MDP is characterized by a tuple of $\left\langle \mathcal{S},\mathcal{A},\mathcal{P},r,\gamma \right\rangle$, where $\mathcal{S}$ is the state space; $\mathcal{A}$ is the action space; $\mathcal{P}:\mathcal{S}\times \mathcal{A}\to \mathcal{S}$ is the state transition function; $r:\mathcal{S}\times \mathcal{A}\to \mathbb{R}$ is the reward function; $\gamma$ is the discount factor. As depicted in Fig.~\ref{/MDP}, at each discrete time step $t$, the agent observes a state ${{\mathbf{s}}_{t}}\in \mathcal{S}$ of the environment, and takes an action ${{\mathbf{a}}_{t}}\in \mathcal{A}$ according to its policy $\pi:\mathcal{S}\to \mathcal{A}$. The environment then gives rise to a scalar reward ${{r}_{t+1}}\in \mathbb{R}$ according to the reward function $r\left( {{\mathbf{s}}_{t}},{{\mathbf{a}}_{t}} \right)$ and evolves to the next state ${{\mathbf{s}}_{t+1}}\in \mathcal{S}$ through state transition function $\mathcal{P}\left( {{\mathbf{s}}_{t}},{{\mathbf{a}}_{t}} \right)$. A complete trajectory of an agent interacting with its environment is known as an episode. This above process is reiterated until the policy convergences. Crucially, the above elements must be reasonably mapped into the GCS process.

\textbf{\textit{State.}} The state space should contain enough information about the subsurface environment. In this study, the state $\mathbf{s}_{t}$ observed by the agent at time step $t$ includes distinct types of data (e.g., dynamic spatial state variables, static spatial system parameters, static vector-type fluid properties), which can be expressed as:
\begin{equation} 
    {{\mathbf{s}}_{t}}=\left({{\mathbf{u}}_{t}}, {{\mathbf{m}}},{{\mathbf{o}}} \right)\in \mathcal{S}
    \label{s}
\end{equation}
where the system parameters $\textbf{m}$ and $\textbf{o}$ are invariant in an episode but vary from different episodes.

\textbf{\textit{Action.}} The action corresponds to the well controls of the injectors and producers. Here, the action $\mathbf{a}_{t}$ executed by the agent at time step $t$ is defined as:
\begin{equation} 
    {{\mathbf{a}}_{t}}=\left( I_{t}^{1},...I_{t}^{i},...,I_{t}^{{{N}_{I}}},P_{t}^{1},...,P_{t}^{j},...,P_{t}^{{{N}_{P}}} \right)\in \mathcal{A}
    \label{a}
\end{equation}
where $I_{t}^{i}$ is the CO$_2$ injection rate of the $i\text{th}$ injector at the time step $t$. $P_{t}^{j}$ is the bottom-hole pressure (BHP) of the $j\text{th}$ producer at the time step $t$. ${{N}_{I}}$ and ${{N}_{P}}$ are the total number of the injectors and producers, respectively.

\textbf{\textit{Reward.}} The reward is a numerical feedback that an agent receives from its environment after taking a certain action in a particular state. The reward function should be designed based on the objective of the optimization problem. Thus, the reward $r_{t}$ received by the agent at time step $t$ is defined as:
\begin{equation} 
    {{r}_{t}}=\left( {{r}_{c}}\cdot {{Q}_{c,t}}-{{r}_{b}}\cdot {{Q}_{b,t}} \right)\cdot \Delta t
    \label{r}
\end{equation}

\textbf{\textit{Policy.}} The policy of the agent is a function that maps current states to actions, which can be implemented by a function approximator with tunable parameters, such as deep neural networks (DNNs). The goal of the DRL agent is to find the optimal policy ${\pi }^{*}$ that maximizes the accumulated discounted reward, i.e.,
\begin{equation} 
    {{\pi }^{*}}=\underset{\pi \in \Pi }{\mathop{\arg \max }}\,{{\mathbb{E}}_{{{\mathbf{s}}_{t}}\sim {{\rho }_{\pi }},{{\mathbf{a}}_{t}}\sim \pi \left( \cdot |{{\mathbf{s}}_{t}} \right)}}\left[ \sum\limits_{t=1}^{H}{{{\gamma }^{t}}}r\left( {{\mathbf{s}}_{t}},{{\mathbf{a}}_{t}} \right) \right]
    \label{pi}
\end{equation}
where $\Pi$ is the set of policies. ${\rho }_{\pi }$ is the state-action visitation distribution of the policy $\pi$.

To derive the optimal policy, existing DRL algorithms entail a large number of calls to the simulator to compute state transitions $\mathcal{P}\left( {{\mathbf{s}}_{t}},{{\mathbf{a}}_{t}} \right)$ and reward values $r\left( {\mathbf{{s}}_{t}},{\mathbf{{a}}_{t}} \right)$, which is computationally expensive. Thus, there is a significant practical interest in developing more efficient algorithms.

\section{Methodology}
\label{sec3}
In this section, we first describe the network architecture and training strategy of the MLD model constructed in this study. Then, we introduce how to incorporate it into the DRL framework and present the workflow of the resulting MSDRL algorithm.

\subsection{Multimodal latent dynamic model}
\begin{figure}[!htb]
    \centering
    \includegraphics[width=0.9\textwidth]{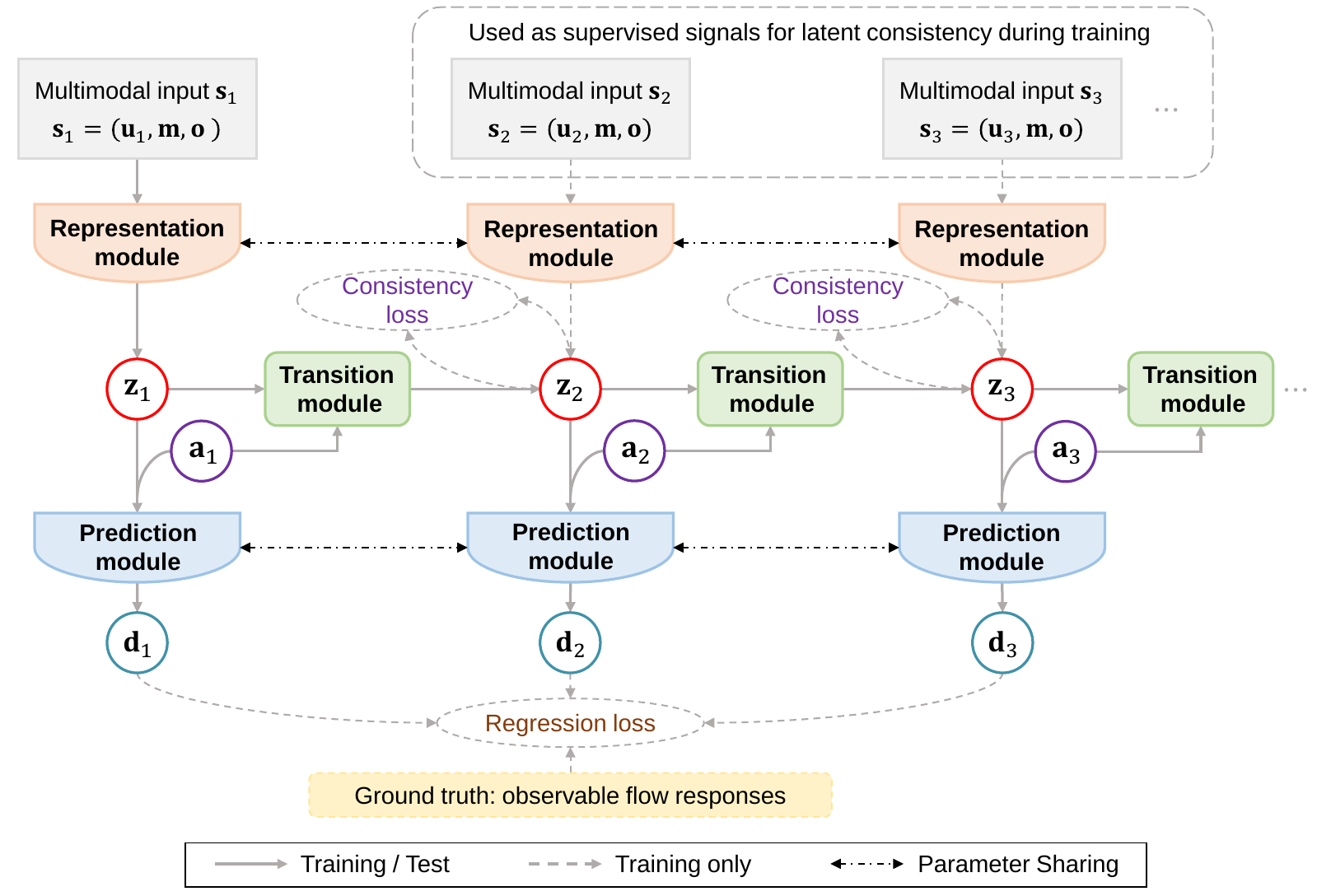}
    \caption{Summary of the MLD model. The MLD model consists of a representation module, a transition module, and a prediction module. The parameters of each module are shared at discrete time steps of an episode to improve efficiency. At the training stage, three modules are jointly optimized by minimizing the regression and consistency loss function. At the test stage (solid gray line), only the first state input is required and the subsequent predictions are made entirely in the latent space.}
    \label{mld}
\end{figure}

To accelerate the policy optimization process, we developed a novel deep learning framework, termed MLD, to approximate the state transitions of the dynamic system and predict the time-varying flow responses. As illustrated in Fig.~\ref{mld}, the MLD model comprises three components: a representation module, a transition module, and a prediction module. 

\subsubsection{Model architecture}

The first component is the representation module ${{R}_{\mathsf{\theta }}}$. At the time step $t$, it receives the multimodal state ${{\mathbf{s}}_{t}}=\left({{\mathbf{u}}_{t}}, {{\mathbf{m}}},{{\mathbf{o}}} \right)\in \mathcal{S}$ as an input and transforms it into a low-dimensional latent state ${{\mathbf{z}}_{t}}\in {{\mathbb{R}}^{{{N}_{z}}}}$, which can be expressed as:
\begin{equation} 
    {{\mathbf{z}}_{t}}={{R}_{\mathsf{\theta }}}\left( {{\mathbf{s}}_{t}} \right)
    \label{rep}
\end{equation}
where $\mathsf{\theta }$ denotes the model parameters. ${{N}_{z}}$ is the dimension of the latent state.

\begin{figure}[!htb]
    \centering
    \includegraphics[width=0.85\textwidth]{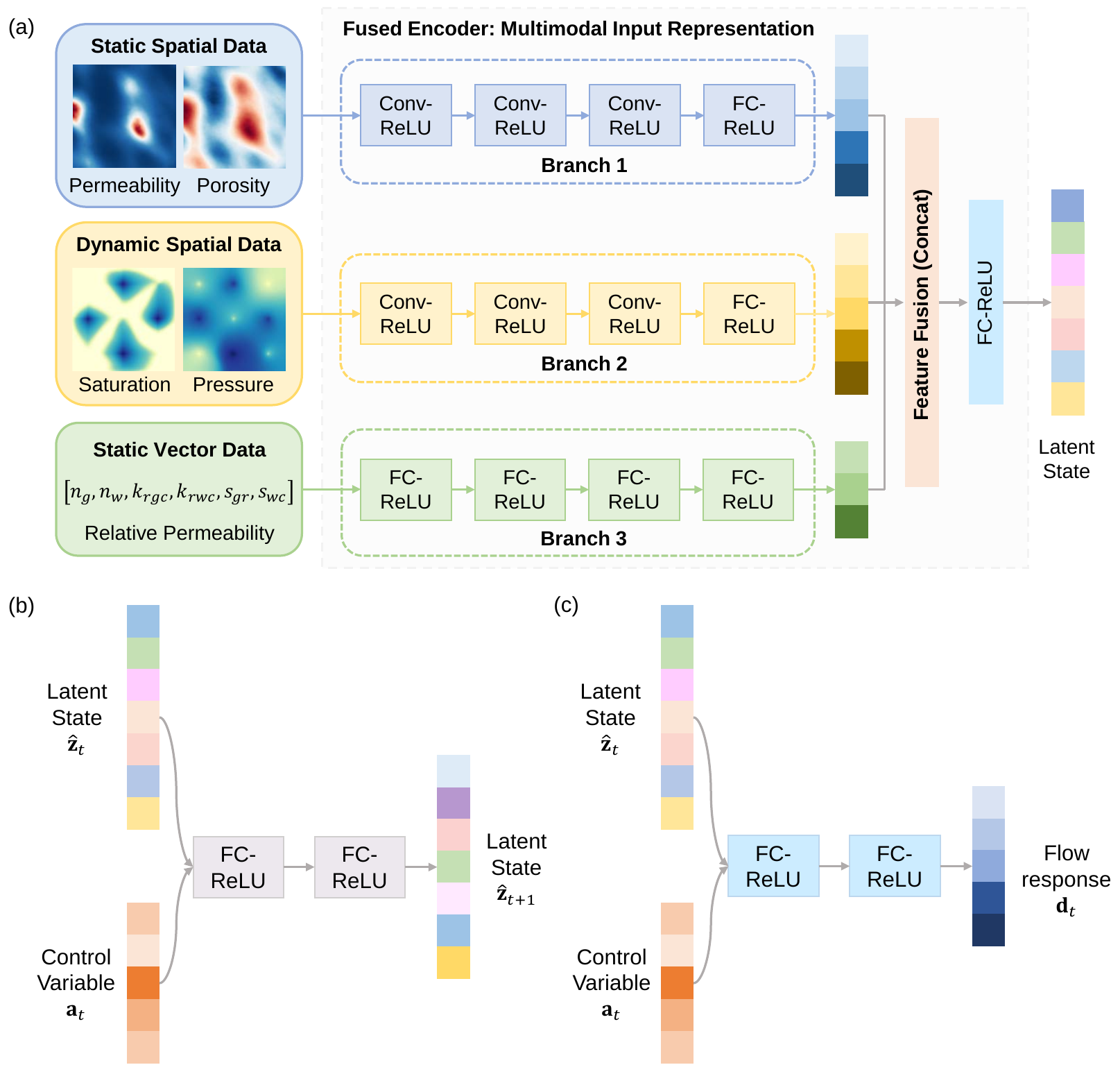}
    \caption{Network architectures of three components. (a) Representation module, implemented as a fused encoder. CNN and MLP are used to extract information from different branches. (b) Transition module, implemented as an MLP. (c) Prediction module, implemented as an MLP. FC and Conv denote the fully connected layer and convolutional layer, respectively. ReLU stands for Rectified Linear Unit, which is a commonly used activation function in deep learning models.}
    \label{mld-archi}
\end{figure}

The model architecture of the representation module is depicted in Fig.~\ref{mld-archi}a. The input includes ${{\mathbf{u}}_{t}}$, (optional) ${{\mathbf{m}}}$, and (optional) ${{\mathbf{o}}}$. When applied to deterministic optimization, only ${{\mathbf{u}}_{t}}$ is required. The addition of ${{\mathbf{m}}}$ and ${{\mathbf{o}}}$ enhances model generality, allowing for generalizable optimization. Without loss of generality, we designed a fused encoder to extract multimodal input features through different branches, such as convolutional neural network (CNN) for spatial data and multilayer perception (MLP) for vector data.

The second component is the transition module ${{T}_{\mathsf{\theta }}}$. At the time step $t$, it takes as input a latent state ${{\mathbf{\hat{z}}}_{t}}\in {{\mathbb{R}}^{{{N}_{z}}}}$ and a control variable ${{\mathbf{a}}_{t}}\in {{\mathbb{R}}^{{{N}_{a}}}}$, and then predicts the latent state ${{\mathbf{\hat{z}}}_{t+1}}\in {{\mathbb{R}}^{{{N}_{z}}}}$ at the next time step $t+1$, which can be expressed as:
\begin{equation} 
    {{\mathbf{\hat{z}}}_{t+1}}={{T}_{\mathsf{\theta }}}\left( {{{\mathbf{\hat{z}}}}_{t}},{{\mathbf{a}}_{t}} \right)
    \label{tran}
\end{equation}
Note that ${{\mathbf{\hat{z}}}_{t+1}}$ differs from ${{\mathbf{z}}_{t+1}}$. The latter is the latent state encoded by the representation module at the time step $t+1$. The model architecture of the transition module is shown in Fig.~\ref{mld-archi}b. Since both ${{\mathbf{\hat{z}}}_{t}}$ and ${{\mathbf{a}}_{t}}$ are one-dimensional vectors, we implemented it as an MLP. Here, our approach inherits a core strength of the previous ROM methods, which can naturally incorporate well controls into the framework \citep{jin2020deep, huang2024application}.
 
The third component is the prediction module ${{P}_{\mathsf{\theta }}}$. At the time step $t$, it takes as input a latent state ${{\mathbf{\hat{z}}}_{t}}\in {{\mathbb{R}}^{{{N}_{z}}}}$ and a control variable ${{\mathbf{a}}_{t}}\in {{\mathbb{R}}^{{{N}_{a}}}}$, and then predicts the flow responses ${{\widehat{\mathbf{d}}}_{t}}\in {{\mathbb{R}}^{{{N}_{d}}}}$, which can be expressed as:
\begin{equation} 
    {{\mathbf{\hat{d}}}_{t}}={{P}_{\mathsf{\theta }}}\left( {{{\mathbf{\hat{z}}}}_{t}},{{\mathbf{a}}_{t}} \right)
    \label{pre}
\end{equation}

The model architecture of the prediction module is shown in Fig.~\ref{mld-archi}c. Similar to the transition module, we found it enough to implement the prediction module as an MLP.

\subsubsection{Training strategy}
The commonly used strategy for model training is to minimize the data mismatch between predicted flow responses ${{\widehat{\mathbf{d}}}_{t}}$ and true values $\mathbf{d}_{t}$. The mean squared error (MSE) loss for the regression problem with $N$ samples can be expressed as:
\begin{equation} 
    {{\mathcal{L}}_{\text{MSE}}}\left( \theta  \right)=\frac{1}{{{N}}H}\sum\limits_{i=1}^{{{N}}}{\sum\limits_{t=1}^{H}{{{l}_{\text{MSE}}}\left( \mathsf{\theta };\mathbf{\hat{d}}_{t}^{i},\mathbf{d}_{t}^{i} \right)}}
    \label{Lm}
\end{equation}
where 
\begin{equation} 
    {{l}_{\text{MSE}}}\left( \mathsf{\theta };\mathbf{\hat{d}}_{t}^{i},\mathbf{d}_{t}^{i} \right)=\left\| {{P}_{\mathsf{\theta }}}\left( \mathbf{\hat{z}}_{t}^{i},\mathbf{a}_{t}^{i} \right)-\mathbf{d}_{t}^{i} \right\|_{2}^{2}
    \label{lmse}
\end{equation}

However, different from standard regression tasks, we are also concerned with state transitions of the dynamic system. Previous methods generally rely on a reconstruction loss to shape latent representations, which suffers from inefficiency and compounding errors, especially for tasks with long time steps. To tackle this issue, we introduced a joint-embedding consistency (JEC) loss for multi-step prediction, which is expressed as:
\begin{equation} 
    {\mathcal L_{\text{JEC}}}\left( \mathsf{\theta } \right)=\frac{1}{NH}\sum\limits_{i=1}^{N}{\sum\limits_{t=1}^{H}{{{l}_{\text{JEC}}}}}\left( \mathsf{\theta };\mathbf{\hat{z}}_{t+1}^{i},\mathbf{z}_{t+1}^{i} \right)
    \label{Lj}
\end{equation}
where 
\begin{equation} 
    {{l}_{\text{JEC}}}\left( \mathsf{\theta };\mathbf{\hat{z}}_{t+1}^{i},\mathbf{z}_{t+1}^{i} \right)=\left\| {{T}_{\mathsf{\theta }}}\left( \mathbf{\hat{z}}_{t}^{i},\mathbf{a}_{t}^{i} \right)-{{R}_{\mathsf{\theta }}}\left( \mathbf{s}_{t+1}^{i} \right) \right\|_{2}^{2}
    \label{ljec}
\end{equation}

At each discrete time step $t$, the JEC loss enforces the future latent state prediction ${{\mathbf{\hat{z}}}_{t+1}}$ to be similar to corresponding ground truth ${{\mathbf{z}}_{t+1}}$. We emphasize that ensuring temporal consistency in the learned latent representations is crucial as it connects all three components of the MLD model. Now, an augmented loss function for joint optimization of three components can be expressed as: 
\begin{equation} 
    \mathcal L\left( \mathsf{\theta } \right)=\lambda {\mathcal L_{\text{MSE}}}\left( \mathsf{\theta } \right)+ {\mathcal L_{\text{JEC}}}\left( \mathsf{\theta } \right)\text{+}\frac{\beta }{\text{2}}\left\| \mathsf{\theta } \right\|_{2}^{2}
    \label{reg}
\end{equation}
where $\lambda $ is the weight coefficient that balances two losses. $\beta$ is known as the weight decay. Taken together, the three terms encourage more accurate long-term predictions. The training procedure of the MLD model is summarized in \textbf{Algorithm~\ref{alg1}}.

\begin{algorithm}
        \small
	\caption{Training procedure of the MLD model}
	\label{alg1}
	\begin{algorithmic}[1]
            \REQUIRE learning rate ${{\eta }_{1}}$, mini-batch size ${{\mathcal{B}}_{1}}$, weight decay $\beta$, weight coefficient $\lambda $
		\STATE \textbf{Initialize}: model parameters $\mathsf{\theta }\leftarrow {{\mathsf{\theta }}_{0}}$
		\FOR{each epoch}
                \FOR{each mini-batch $\left\{ \left( \mathbf{s}_{t}^{i},\mathbf{a}_{t}^{i},\mathbf{d}_{t}^{i},\mathbf{s}_{t+1}^{i} \right)_{t=1}^{H} \right\}_{i=1}^{{{\mathcal{B}}_{1}}}$ of the training set}
		          \STATE $\mathbf{\hat{z}}_{1}^{1:{{\mathcal{B}}_{1}}}=\mathbf{z}_{1}^{1:{{\mathcal{B}}_{1}}}={{R}_{\mathsf{\theta }}}\left( \mathbf{s}_{1}^{1:{{\mathcal{B}}_{1}}} \right)$
                    \STATE ${\mathcal L_{\text{MSE}}}=0$, ${\mathcal L_{\text{JEC}}}=0$
                    \FOR{$t=1,...,H$}
                        \STATE $\mathbf{\hat{d}}_{t}^{1:{{\mathcal{B}}_{1}}}={{P}_{\mathsf{\theta }}}\left( \mathbf{\hat{z}}_{t}^{1:{{\mathcal{B}}_{1}}},\mathbf{a}_{t}^{1:{{\mathcal{B}}_{1}}} \right)$  
                        \STATE $\mathbf{\hat{z}}_{t+1}^{1:{{\mathcal{B}}_{1}}}={{T}_{\mathsf{\theta }}}\left( \mathbf{\hat{z}}_{t}^{1:{{\mathcal{B}}_{1}}},\mathbf{a}_{t}^{1:{{\mathcal{B}}_{1}}} \right)$ 
                        \STATE $\mathbf{z}_{t+1}^{1:{{\mathcal{B}}_{1}}}={{R}_{\mathsf{\theta }}}\left( \mathbf{s}_{t+1}^{1:{{\mathcal{B}}_{1}}} \right)$ 
                        \STATE ${\mathcal L_{\text{MSE}}}\leftarrow {\mathcal L_{\text{MSE}}}+\mathrm{}\sum\nolimits_{i=1}^{{{\mathcal{B}}_{1}}}{{{l}_{\text{MSE}}}\left( \mathsf{\theta };\mathbf{\hat{d}}_{t}^{i},\mathbf{d}_{t}^{i} \right)}$ 
                        \STATE ${\mathcal L_{\text{JEC}}}\leftarrow {\mathcal L_{\text{JEC}}}+\mathrm{}\sum\nolimits_{i=1}^{{{\mathcal{B}}_{1}}}{{{l}_{\text{JEC}}}\left( \mathsf{\theta };\mathbf{\hat{z}}_{t+1}^{i},\mathbf{z}_{t+1}^{i} \right)}$ 
                    \ENDFOR
                    \STATE $\mathcal L=\lambda {\mathcal L_{\text{MSE}}}+{\mathcal L_{\text{JEC}}}+\frac{\beta }{2}\left\| \theta  \right\|_{2}^{2}$
                    \STATE \ $\mathsf{\theta }\leftarrow \mathsf{\theta }-{{\eta }_{1}}{{\nabla }_{\mathsf{\theta }}}\mathcal L$
                \ENDFOR
            \ENDFOR
		\ENSURE $\mathsf{\theta }$
	\end{algorithmic}  
\end{algorithm}

After the offline training stage, the MLD model can be used as an alternative to numerical simulation (described by Eq.~\ref{stateeq} and Eq.~\ref{outputeq}) for forward flow prediction, as illustrated by the solid gray line in Fig.~\ref{mld}. It can be seen that during inference, only the first raw state ${{\mathbf{s}}_{1}}$ is encoded using ${{R}_{\mathsf{\theta }}}\left( {{\mathbf{s}}_{1}} \right)$ and subsequent recurrent predictions are made entirely in the latent space with: 
\begin{equation} 
    {{\mathbf{z}}_{1}}={{R}_{\mathsf{\theta }}}\left( {{\mathbf{s}}_{1}} \right),{{\mathbf{\hat{z}}}_{2}}={{T}_{\mathsf{\theta }}}\left( {{\mathbf{z}}_{1}},{{\mathbf{a}}_{1}} \right),...,{{\mathbf{\hat{z}}}_{H+1}}={{T}_{\mathsf{\theta }}}\left( {{\mathbf{\hat{z}}}_{H}},{{\mathbf{a}}_{H}} \right)
    \label{flow}
\end{equation}

\subsection{Soft actor-critic}
SAC is a state-of-the-art DRL algorithm and has been widely applied to address continuous optimization tasks \citep{haarnoja2018soft, haarnoja2018soft2, erdman2022identifying}. In this study, SAC is employed to solve the sequential well control optimization problems. SAC utilizes a maximum entropy mechanism and optimizes the policy that maximizes both the long-term expected reward and the entropy of the policy. In this setting, the optimal policy ${{\pi }^{*}}$ is given by:
\begin{equation}
    {{\pi }^{*}}=\underset{\pi \in \Pi }{\mathop{\arg \max }}\,\text{ }{{\mathbb{E}}_{{{\mathbf{s}}_{t}}\sim {{\rho }_{\pi }},{{\mathbf{a}}_{t}}\sim \pi \left( \cdot |{{\mathbf{s}}_{t}} \right)}}\left[ \sum\limits_{t=1}^{H}{{{\gamma }^{t}}}\left( r\left( {{\mathbf{s}}_{t}},{{\mathbf{a}}_{t}} \right)+\alpha \mathcal{H}\left( \pi \left( \cdot |{{\mathbf{s}}_{t}} \right) \right) \right) \right]
    \label{pi*}
\end{equation}
where $\mathcal{H}$ is the entropy term. $\alpha $ is the temperature coefficient balancing exploration and exploitation. As we will see later, $\alpha $ has a significant impact on the performance of the algorithm.

\begin{figure}[!htb]
    \centering
    \includegraphics[width=0.7\textwidth]{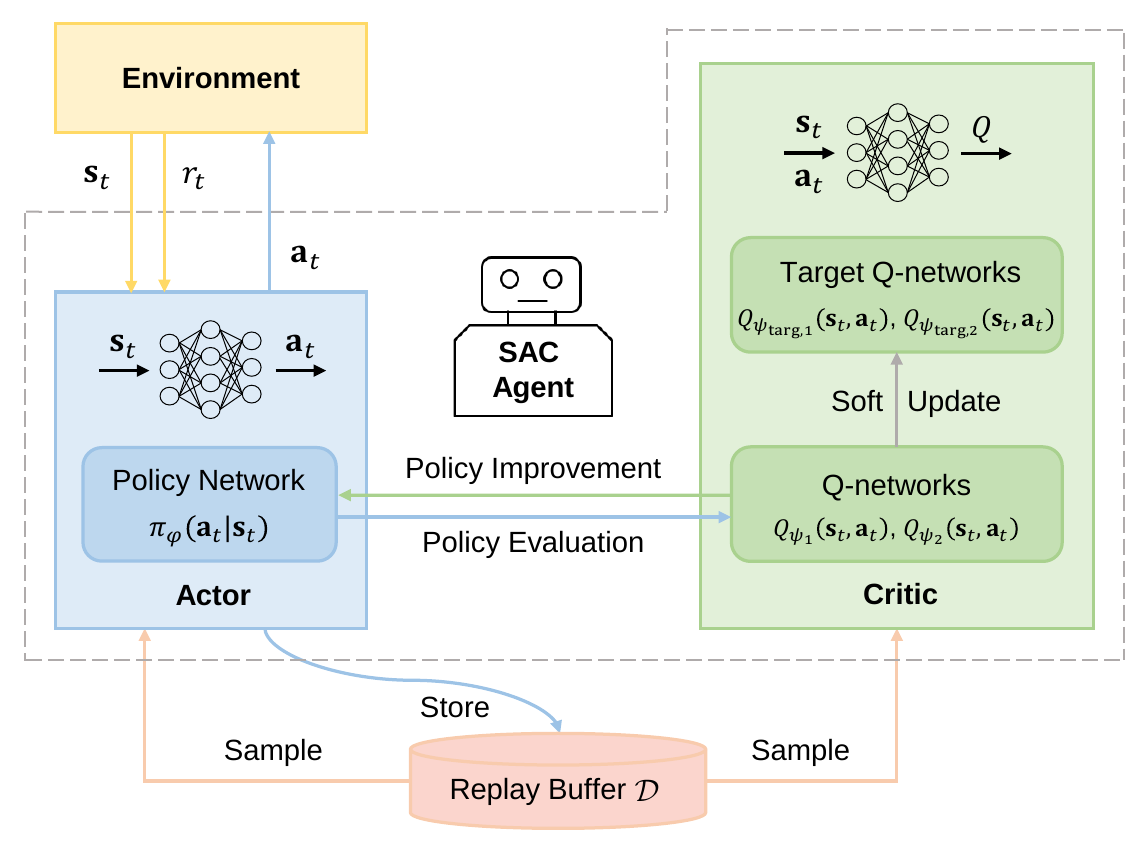}
    \caption{Structure of the SAC agent. SAC agent contains an actor and a critic. For the actor, the input is the state and the output is the action. For the critic, the input is the state-action pair and the output is the Q-value.}
    \label{sac}
\end{figure}

The schematic of the SAC-based learning process is shown in Fig.~\ref{sac}. To treat high-dimensional state space and action space, DNNs are used to approximate the policy function and the Q-function. Specifically, the policy network ${{\pi }_{\mathsf{\varphi }}}\left( {{\mathbf{a}}_{t}}|{{\mathbf{s}}_{t}} \right)$ parameterized by $\mathsf{\varphi }$ plays a role of "actor" that takes an action by observing the current state. The Q-network ${Q}_{{{\mathsf{\psi }}}}\left( {{\mathbf{s}}_{t}},{{\mathbf{a}}_{t}} \right)$ parameterized by ${{\mathsf{\psi }}}$ plays a role of "critic" that judges the action made by the actor, thus providing feedback to adjust and improve the actor’s decision. Two Q-networks that share the same structure are employed to mitigate positive bias. In addition, the target Q-networks are used to stabilize the training process. During the optimization process, the SAC agent stores the experience $\left( {{\mathbf{s}}_{t}},{{\mathbf{a}}_{t}},{{r}_{t}},{{\mathbf{s}}_{t+1}} \right)$ gained from interacting with the environment in a replay buffer $\mathcal{D}$. Through sampling a mini-batch of tuples from $\mathcal{D}$, the parameters of the actor and critic can be updated using backpropagation. The detailed training process can be found in ~\ref{apa}.

\subsection{Workflow of MSDRL algorithm}
\begin{figure}[!htb]
    \centering
    \includegraphics[width=\textwidth]{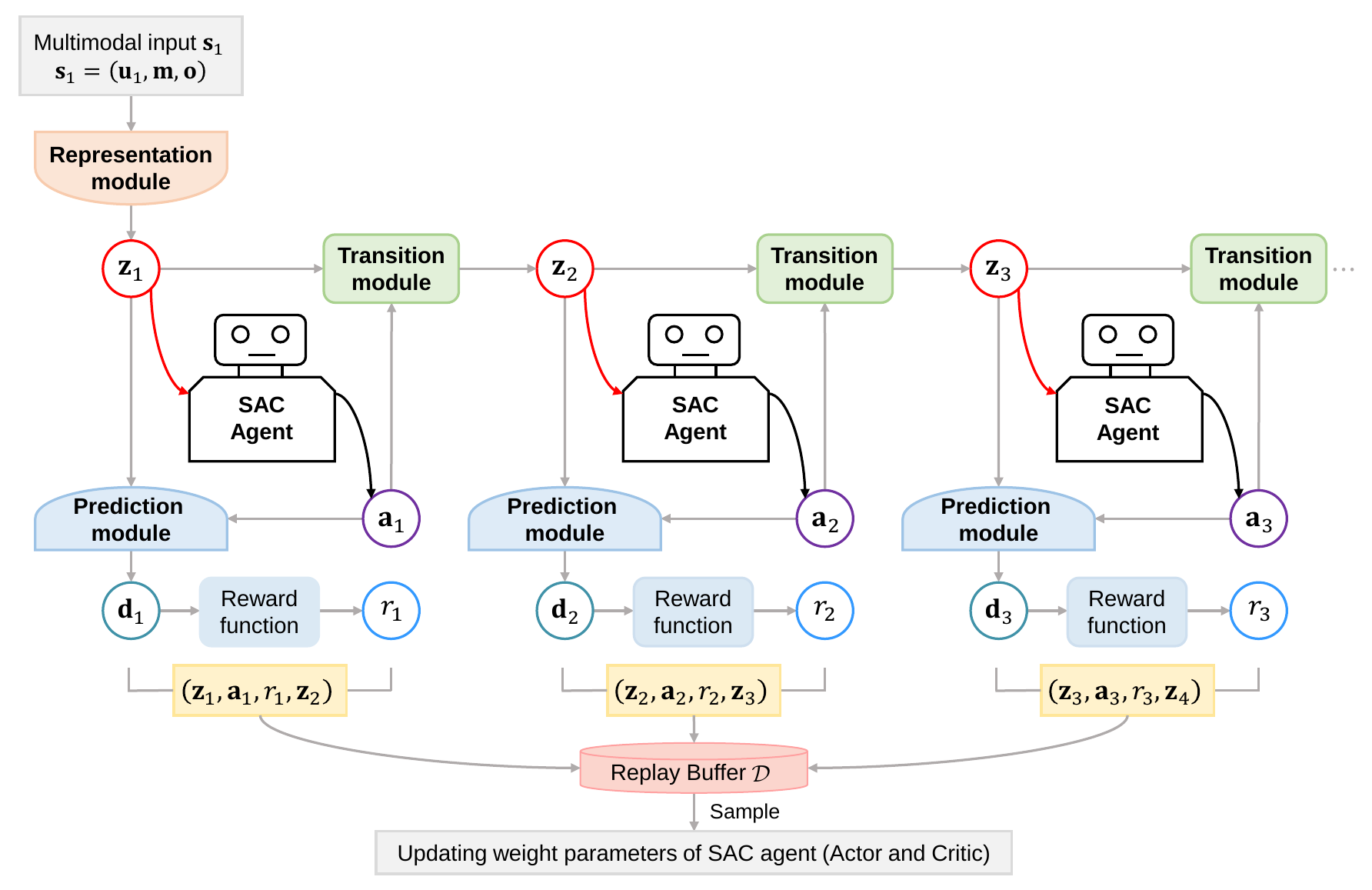}
    \caption{Workflow of the MSDRL algorithm. The learned MLD model serves as an interactive environment to train a SAC agent that learns control policies entirely from the predictions in the compact latent space.}
    \label{msdrl}
\end{figure}

Based on the constructed MLD model and SAC agent model, we proposed a data-driven optimization algorithm, termed MSDRL. During the policy optimization process, the learned MLD model serves as an interactive environment to provide fast feedback, enabling the SAC agent to explore a large search space with small computational costs. Regarding that it is extremely hard to train the agent based on the original multimodal state space, MSDRL leverages the learned representation of the multimodal state and trains the agent fully in the compact latent space. The overall workflow is illustrated in Fig.~\ref{msdrl}. Specifically, at the initial time step, given a multimodal state ${{\mathbf{s}}_{1}}$, the representation module encodes it to a low-dimensional latent state ${{\mathbf{z}}_{1}}$ and provides it to SAC agent. The SAC agent executes an action ${{\mathbf{a}}_{1}}$ according to its current policy ${{\pi }_{\mathsf{\varphi }}}\left( {{\mathbf{a}}_{1}}|{{\mathbf{z}}_{1}} \right)$. The prediction module receives ${{\mathbf{z}}_{1}}$ and ${{\mathbf{a}}_{1}}$, and forecasts the flow responses ${{\mathbf{d}}_{1}}$, which is then used to calculate a reward ${{r}_{1}}$ with the reward function Eq.~\ref{r}. At the same time, the transition module evolves the system state from ${{\mathbf{z}}_{1}}$ to ${{\mathbf{z}}_{\text{2}}}$ in the latent space. The episode (i.e., a forward process) terminates when it reaches the time step $t=H$. The experience tuples $\left( {{\mathbf{z}}_{t}},{{\mathbf{a}}_{t}},{{r}_{t}},{{\mathbf{z}}_{t+1}} \right)_{t=1}^{H}$ are appended into a replay buffer $\mathcal{D}$. While interacting with the environment, the SAC agent randomly samples mini-batch data from $\mathcal{D}$ to update its weight parameters. The above process is repeatedly performed and the SAC agent eventually learns the optimal policy. The pseudocode of MSDRL is provided in \textbf{Algorithm~\ref{alg2}}.

\begin{algorithm}[!htb]
        \small
	\renewcommand{\algorithmicrequire}{\textbf{Input:}}
	\renewcommand{\algorithmicensure}{\textbf{Output:}}
	\caption{Pseudocode of MSDRL}
	\label{alg2}
	\begin{algorithmic}[1]
            \REQUIRE learned MLD model parameters $\mathsf{\theta }$, learning rate ${{\eta }_{\text{2}}}$, mini-batch size ${{\mathcal{B}}_{\text{2}}}$
		\STATE \textbf{Initialize}: policy ${{\pi }_{\mathsf{\varphi }}}\left( {{\mathbf{a}}_{t}}|{{\mathbf{s}}_{t}} \right)$ with parameters $\mathsf{\varphi }$, two Q-networks ${{Q}_{{{\mathsf{\psi }}_{1}}}}\left( {{\mathbf{s}}_{t}},{{\mathbf{a}}_{t}} \right)$ and ${{Q}_{{{\mathsf{\psi }}_{2}}}}\left( {{\mathbf{s}}_{t}},{{\mathbf{a}}_{t}} \right)$ with parameters ${{\mathsf{\psi }}_{\text{1}}}$ and ${{\mathsf{\psi }}_{2}}$, two target Q-networks with parameters ${{\mathsf{\psi }}_{\text{targ,}1}}\leftarrow {{\mathsf{\psi }}_{1}}$ and ${{\mathsf{\psi }}_{\text{targ},2}}\leftarrow {{\mathsf{\psi }}_{2}}$, a replay memory $\mathcal{D}\leftarrow \varnothing $
		\FOR{each iteration}
                \STATE Observe initial state ${{\mathbf{s}}_{1}}$
                \STATE \textit{/* Interact with the MLD model to collect data */}
                \STATE Obtain latent state ${{\mathbf{z}}_{1}}={{R}_{\mathsf{\theta }}}\left( {{\mathbf{s}}_{1}} \right)$
                \FOR{time step $t=1,...H$}
                    \STATE Take action ${{\mathbf{a}}_{t}}={{\pi }_{\mathsf{\varphi }}}\left( {{\mathbf{a}}_{t}}|{{{\mathbf{z}}}_{t}} \right)$
                    \STATE Obtain next latent state ${{\mathbf{z}}_{t+1}}={{T}_{\mathsf{\theta }}}\left( {{{\mathbf{z}}}_{t}},{{\mathbf{a}}_{t}} \right)$ and reward ${{r}_{t}}=\mathcal{R}\left( {{{\mathbf{d}}}_{t}} \right)=\mathcal{R}\left( {{P}_{\mathsf{\theta }}}\left( {{{\mathbf{z}}}_{t}},{{\mathbf{a}}_{t}} \right) \right)$
                    \STATE Store the experience tuple into the replay buffer $\mathcal{D}\leftarrow \mathcal{D}\cup \left\{ \left( {{{\mathbf{z}}}_{t}},{{\mathbf{a}}_{t}},{{r}_{t}},{{\mathbf{z}}_{t+1}} \right) \right\}$
                \ENDFOR
                \STATE \textit{/* Train the neural networks using the experience */}
                \FOR{each training step}
                    \STATE Sample a mini-batch of tuple $\left\{ {{\left( {{\mathbf{z}}_{i}},{{\mathbf{a}}_{i}},{{r}_{i}},{{\mathbf{z}}_{i+1}} \right)}} \right\}_{i=1}^{{{\mathcal{B}}_{2}}}$ from $\mathcal{D}$
                    \STATE Update Q-functions parameters ${{\mathsf{\psi }}_{j}}\leftarrow {{\mathsf{\psi }}_{j}}-{{\eta }_{2}}{{\nabla }_{{{\mathsf{\psi }}_{j}}}}{\mathcal L_{Q}}\left( {{\mathsf{\psi }}_{j}} \right),\text{for }j\in \left\{ 1,2 \right\}$
                    \STATE Update policy parameters $\mathsf{\varphi }\leftarrow \mathsf{\varphi }-{{\eta }_{2}}{{\nabla }_{\mathsf{\varphi }}}{\mathcal L_{\pi }}\left( \mathsf{\varphi } \right)$
                    \STATE Adjust temperature parameter $\alpha \leftarrow \alpha -{{\eta }_{2}}{{\nabla }_{\alpha }}\mathcal L\left( \alpha  \right)$
                    \STATE Update target Q-functions parameters ${{\mathsf{\psi }}_{\text{targ},j}}\leftarrow \tau {{\mathsf{\psi }}_{j}}+(1-\tau ){{\mathsf{\psi }}_{\text{targ},j}},\text{for }j\in \left\{ 1,2 \right\}$
                \ENDFOR
            \ENDFOR
		\ENSURE $\mathsf{\varphi }$, ${{\mathsf{\psi }}_{1}}$, ${{\mathsf{\psi }}_{2}}$
	\end{algorithmic}  
\end{algorithm} 

\section{Case studies}
\label{sec4}

In this section, the proposed MSDRL algorithm was employed for the optimization of CO$_2$ storage in a heterogeneous saline aquifer. The optimization efficiency and generalization ability of MSDRL were demonstrated through different optimization scenarios. Without loss of generality, all numerical simulation runs were performed using high-fidelity simulator ECLIPSE \citep{pettersen2006basics}.

\subsection{Case1: deterministic optimization scenario}
\begin{figure}[!htb]
    \centering
    \includegraphics[width=0.95\textwidth]{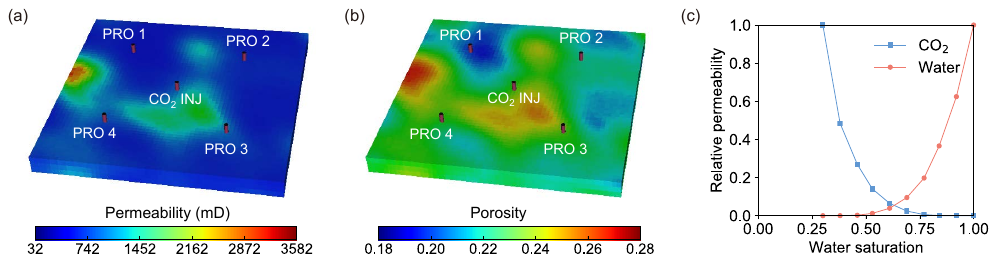}
    \caption{Description of aquifer model information. (a) Permeability field. (b) Porosity field. (c) Relative permeability curve.}
    \label{reservoir}
\end{figure}

The well control optimization was first carried out for a given 3D aquifer model, in which the model properties (permeability field, porosity field, and relative permeability) are assumed to be known, as shown in Fig.~\ref{reservoir}. The aquifer model has 60×60×3 grid blocks. The lateral grid resolution is 60 m by 60 m, and the layer thickness is 7 m. The datum depth is 1411.7 m and the initial pressure at the datum depth is 175.16 bar. All boundaries are assigned as no-flow boundaries. The supercritical CO$_2$ is injected at the center location and the brine is extracted from four producers over 20 years. The CO$_2$ injector is operated following specified rates. The allowable upper and lower bounds are 1.5×10$^6$ m$^3$/day and 5.0×10$^5$ m$^3$/day, respectively. The producers are controlled by BHP. The allowable upper and lower bounds are 170 bar and 150 bar, respectively. The well control operations are changed every 1 year, so the dimension of the optimization problem is (1+4)×20=100. This case was used to test the optimization effectiveness and efficiency of MSDRL.

\subsubsection{Evaluation of MLD model performance}
To construct the MLD model for optimization, 500 time-varying well control schemes within the given range were generated by Latin hypercube sampling, followed by 500 sets of samples obtained by running the simulator. Two groups of well controls randomly selected from the dataset are shown in Fig.~\ref{wc}. With reference to Eq.~\ref{reg}, the MLD model was trained to minimize the augmented loss function $\mathcal L$. The hyperparameter settings during the training process are provided in Table~\ref{tableS1} in ~\ref{apb}. Two commonly used metrics, coefficient of determination R${^2}$ and root-mean-square error RMSE, were introduced to evaluate the performance of the MLD model quantitatively, which are calculated as Eq.~\ref{r2v} and Eq.~\ref{rmsev}, respectively:
\begin{figure}[htb]\centering
    \includegraphics[width=0.95\textwidth]{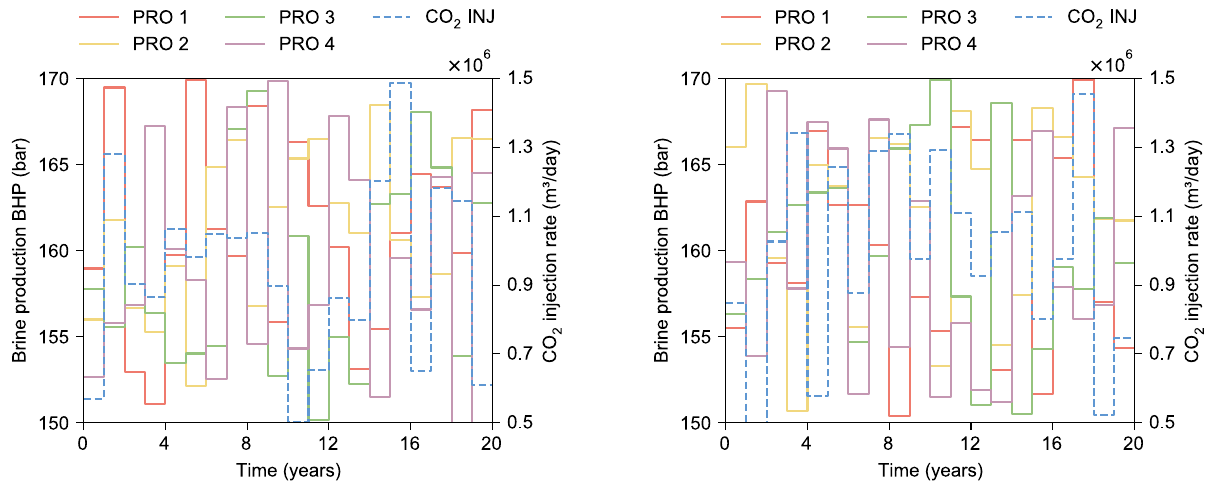}
    \caption{Two groups of randomly selected well control sequences from the dataset.}
    \label{wc}
\end{figure}
\begin{equation}
    {\rm R{^2}}=1-\frac{\sum\nolimits_{i=1}^{N_s}{\sum\nolimits_{t=1}^{H}{\left\| \mathbf{\hat{d}}_{t}^{i}-\mathbf{d}_{t}^{i} \right\|}_{2}^{2}}}{\sum\nolimits_{i=1}^{N_s}{\sum\nolimits_{t=1}^{H}{\left\| {{{\mathbf{\bar{d}}}}_{t}}-\mathbf{d}_{t}^{i} \right\|}_{2}^{2}}}
    \label{r2v}
\end{equation}
\begin{equation}
    {\rm RMSE}=\sqrt{\frac{1}{N_s}\sum\nolimits_{i=1}^{N_s}{\sum\nolimits_{t=\text{1}}^{H}{\left\| \mathbf{\hat{d}}_{t}^{i}-\mathbf{d}_{t}^{i} \right\|}_{2}^{2}}}
    \label{rmsev}
\end{equation}
where $N_s$ is the number of samples. ${{\mathbf{d}}_{t}^{i}}$ and ${{\mathbf{\hat{d}}}_{t}^{i}}$ denote the simulation result and prediction result, respectively. ${{\mathbf{\bar{d}}}_{i}}$ denotes the mean value of the simulation results. A higher R${^2}$-score value (closer to 1) and a lower RMSE value indicate a better model performance.

\begin{figure}[htb]\centering
    \includegraphics[width=\textwidth]{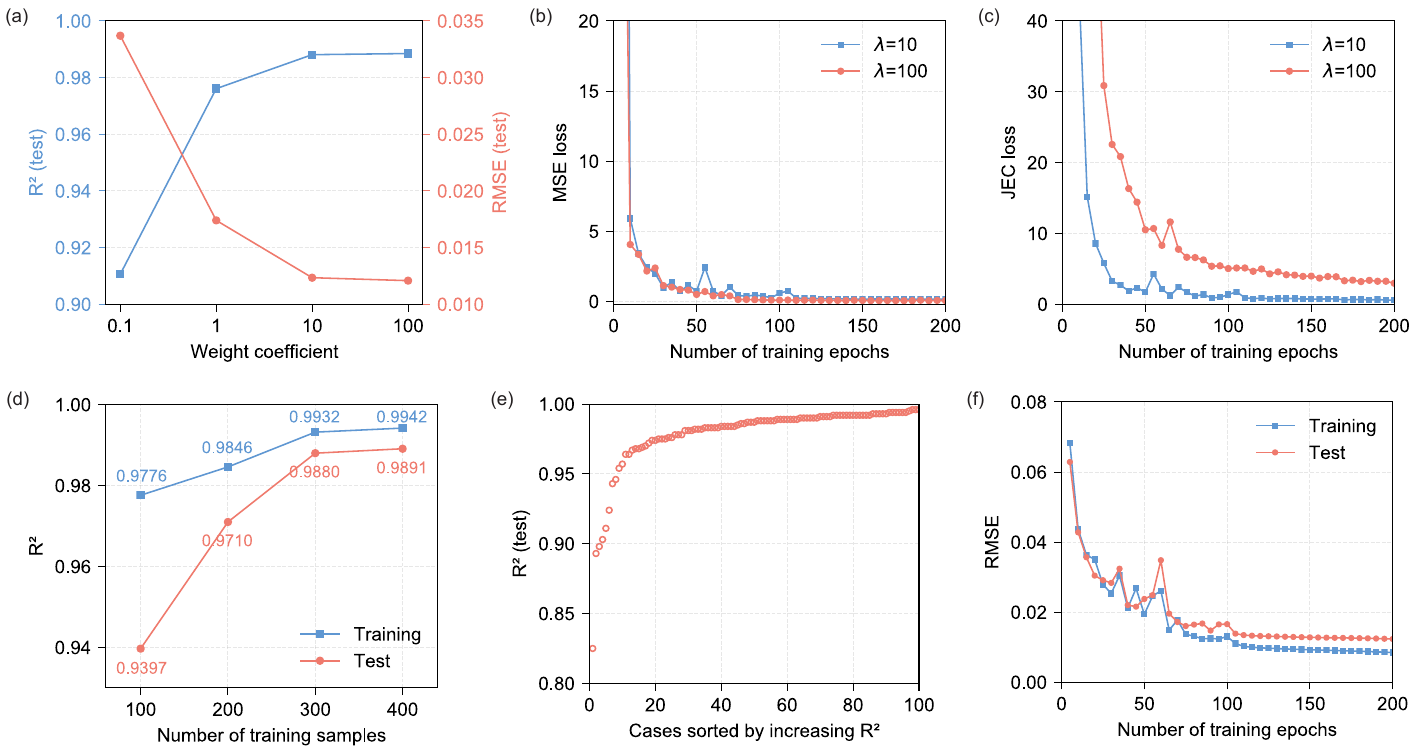}
    \caption{Performance evaluation results of the MLD model. (a) R${^2}$ and RMSE results of models with different $\lambda$ values evaluated on 100 test samples. (b)-(c) Evolution of the MSE loss and JEC loss during the training process. (d) R${^2}$ results of models trained with different training sample sizes. (e) R${^2}$ scores, sorted in increasing order, for all 100 test cases. (f) Evolution of RMSE with the number of epochs during the training process.}
    \label{assess1}
\end{figure}

The weight coefficient $\lambda$ decides the proportion of the regression loss in the total loss. To investigate the impact of $\lambda$ on the prediction accuracy of the MLD model, we tested four different values of $\lambda$ ranging from 0.1 to 100. In these experiments, the training sample size $N_{train}$ was set to 300. Fig.~\ref{assess1}a shows the R${^2}$ and RMSE results of models with different $\lambda$ values evaluated on 100 test samples. It can be seen that among the four models, the models with $\lambda$ = 10 and 100 have better performance. The decays of the MSE loss and JEC loss during the training process are depicted in Fig.~\ref{assess1}b and Fig.~\ref{assess1}c, respectively. It is observed that the model with $\lambda$ = 100 focuses more on flow response regression (smaller MSE loss) and exhibits relatively large errors in approximating the evolution of the system state (larger JEC loss). In contrast, the model with $\lambda$ = 10 is able to better balance these two losses, achieving good convergence for both quantities. Thus, we specified $\lambda$ = 10 for cases considered in this study.

We further assessed the influence of different training sample sizes on the performance of the MLD model. We used 100, 200, 300, and 400 samples as four training sets and the additional 100 samples as the test set. The evaluation results of R${^2}$ with different training sample sizes are depicted in Fig.~\ref{assess1}d. It is observed that as more training samples are available, the predictive accuracy of the MLD model increases. With 300 training samples, the model achieved a high R${^2}$ score of 0.9880 for the test set. Fig.~\ref{assess1}e displays the R${^2}$ scores, sorted in increasing order, for all 100 test cases. The P$_{10}$, P$_{50}$, and P$_{90}$ of test R${^2}$ here are 0.9572, 0.9874, and 0.9933, respectively. The evolution of RMSE with the number of training epochs for $N_{train}$=300 is shown in Fig.~\ref{assess1}f. It can be seen that RMSE begins to converge after 100 epochs and the learning curve is quite smooth, suggesting that the MLD model can learn nonlinear flow dynamics with strong stability. The above results indicate that the MLD model trained with 300 samples can be used as a reliable alternative to the simulator for providing accurate flow predictions.

\subsubsection{Analysis of optimization results}

The trained MLD model was then integrated into the SAC framework to optimize well control policies. We referred to prior research \citep{kawata2017some} to set the economic parameters. Specifically, the brine treatment cost was set to 10 USD/m$^3$. The capture and storage cost of CO$_2$ was set to 35 USD/ton. The CO$_2$ tax was set to 50 USD/ton. Therefore, the profit from CO$_2$ injection is 15 USD/ton (0.0246 USD/m$^3$). Note that if these economic parameters are modified, the MLD model does not need to be retrained as it can provide well-by-well rates as a function of time rather than a single NPV value. The hyperparameter settings for SAC are presented in Table~\ref{tableS2} in ~\ref{apb}.

\begin{figure}[!htb]
    \centering
    \includegraphics[width=0.97\textwidth]{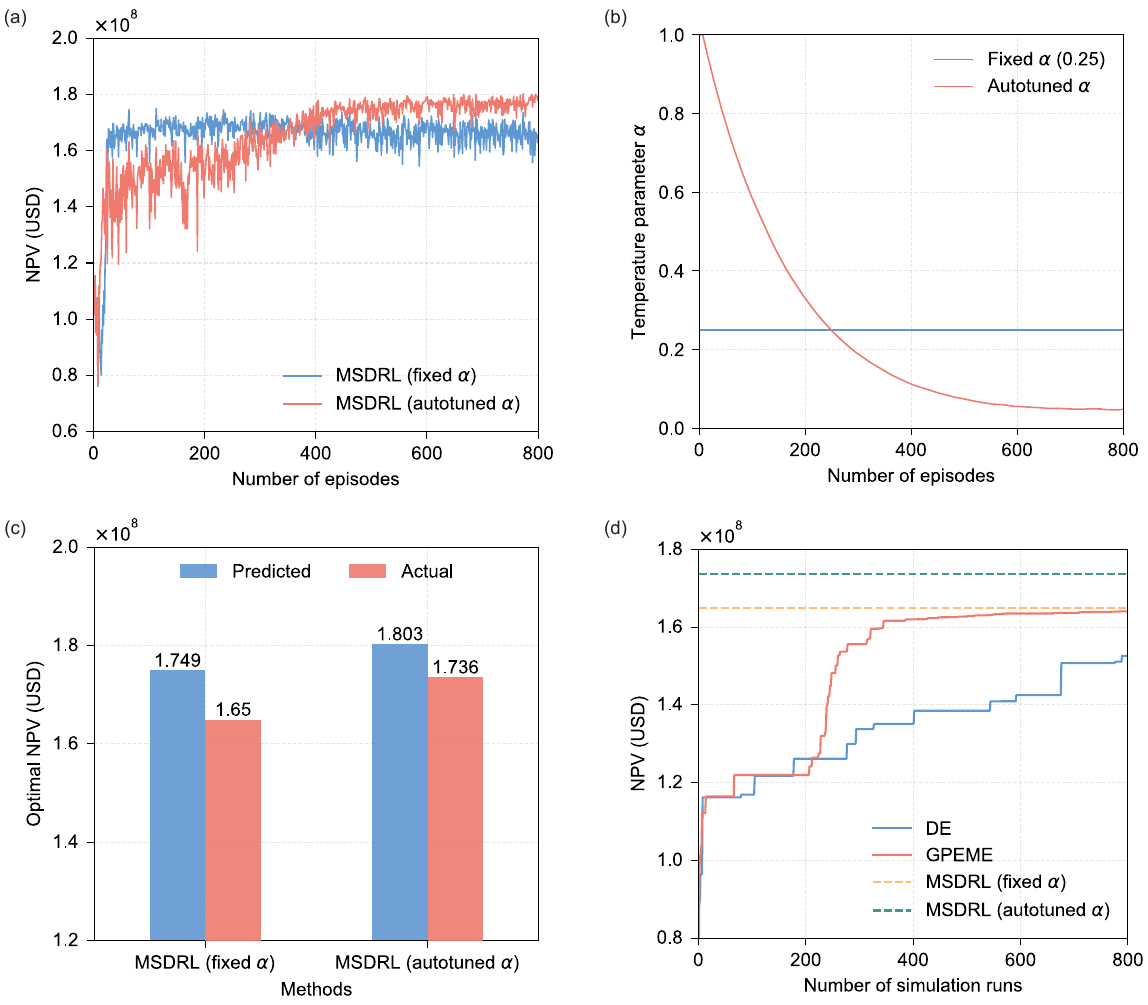}
    \caption{Analysis and comparison of optimization results. (a) Evolution of NPV with the number of episodes during the policy optimization process. (b) Evolution of $\alpha$ with the number of episodes during the policy optimization process. (c) Comparison of predicted optimal NPV values and actual NPV values. (d) Convergence curves of DE and GPEME, along with the performance comparison of different algorithms.}
    \label{opt1}
\end{figure}

As noted earlier, the temperature coefficient $\alpha$ is a key parameter involved in the MSDRL algorithm. To examine its impact on the optimization performance, we trained the agent with two strategies, which are denoted as MSDRL (fixed $\alpha$) and MSDRL (autotuned $\alpha$), respectively. For demonstration purposes, we specified $\alpha$=0.25 for MSDRL (fixed $\alpha$). The evolution of NPV with the number of episodes during the optimization process is shown in Fig.~\ref{opt1}a. It is observed that the NPV obtained by the agent gradually rises and becomes stable as the number of interactions with the MLD model increases. This indicates that the agent is capable of utilizing the accumulated experience to improve its policy. MSDRL (fixed $\alpha$) converged quickly in the early stage, but insufficient exploration of the environment induced the policy to get trapped in local optima. As a comparison, by dynamically adjusting $\alpha$ during the policy optimization process (Fig.~\ref{opt1}b), MSDRL (autotuned $\alpha$) achieved an improved trade-off between exploration and exploitation and ultimately obtained better optimization results. For this case, the optimal NPV values predicted by MSDRL (fixed $\alpha$) and MSDRL (autotuned $\alpha$) are 1.749×10$^8$ USD and 1.803×10$^8$ USD, respectively. Considering that the approximation error of the MLD model may affect the accuracy of the optimization results, we entered the optimal well control schemes into the simulator to determine the actual NPV values. As shown in Fig.~\ref{opt1}c, the corresponding results for MSDRL (fixed $\alpha$) and MSDRL (autotuned $\alpha$) are 1.650×10$^8$ USD and 1.736×10$^8$ USD, respectively, which are approximately 3.7$\sim $5.7\% less than their predicted values. This high consistency further proves the reliability of the MLD model in the policy optimization process.

We also performed the optimization process with differential evolution algorithm (DE) \citep{das2010differential} and Gaussian process surrogate-assisted evolutionary algorithm (GPEME) \citep{liu2013gaussian}, which have been widely used in subsurface energy system design. The convergence curves of DE and GPEME, along with the performance comparison of different algorithms are shown in Fig.~\ref{opt1}d. It is evident that MSDRL (autotuned $\alpha$) attains the highest NPV, indicating that it discovers superior well control schemes compared to those of other evaluated algorithms. Additionally, DE had the worst result and failed to search for a satisfactory solution. GPEME enhanced the overall performance by introducing a surrogate-aware search mechanism, obtaining results comparable to that of MSDRL (fixed $\alpha$).

Table~\ref{table1} summarizes the computational requirements associated with each method. As MSDRL (fixed $\alpha$) and MSDRL (autotuned $\alpha$) have the same computational costs, we uniformly denote them as MSDRL. In our evaluations, MSDRL performed 300 simulation runs to construct the dataset in the offline stage, and the model training required 4.49 minutes on a single GTX 1080 GPU. Once trained, it sped up online flow evaluation more than 10$^4$ times to the full-order compositional simulation. Thus, the time consumption for model training and evaluation can be ignored since it is insignificant compared to the runtime of numerical simulations. By comparison, DE and GPEME required 800 simulation runs during the optimization process, which takes more than 50 hours. Overall, MSDRL reduces the computational costs by more than 60\%.

\begin{table}[!htb]
\small
\centering
\caption{Comparison of computational requirements associated with each method.}
\label{table1}
\begin{tabular}{cccc}
\hline  
\begin{tabular}[c]{@{}c@{}}Optimization\\ Method\end{tabular} & \begin{tabular}[c]{@{}c@{}}Number of simulation \\ evaluations\end{tabular} & \begin{tabular}[c]{@{}c@{}}Number of surrogate \\ evaluations\end{tabular} & \begin{tabular}[c]{@{}c@{}}Total runtime \\ (hours)\end{tabular} \\ \hline
DE                                                            & 800                                                                         & 0                                                                          & 50.268                                                           \\
GPEME                                                         & 800                                                                         & 30000                                                                      & 53.464                                                           \\
MSDRL                                                         & 300                                                                         & 800                                                                        & 19.073                                                           \\ \hline
\end{tabular}
\end{table}

\begin{figure}[!htb]
    \centering
    \includegraphics[width=\textwidth]{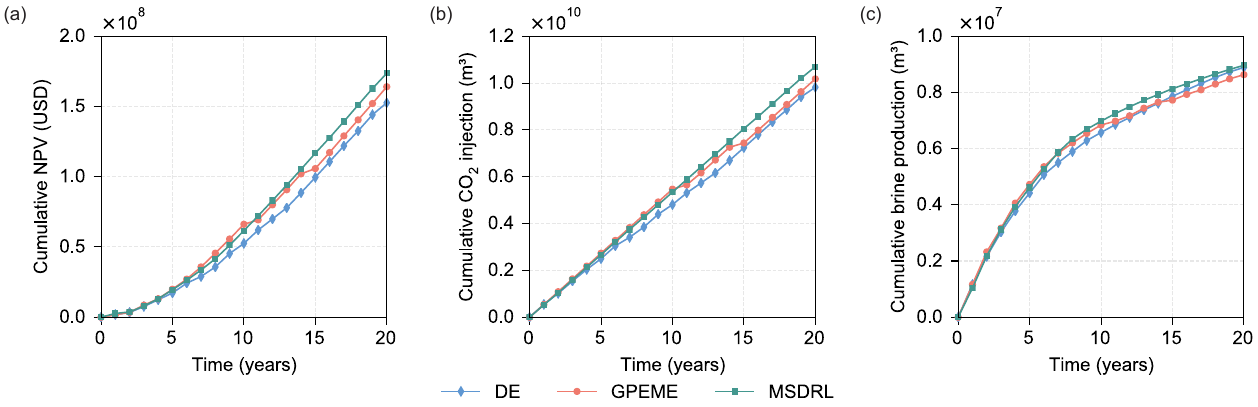}
    \caption{Simulation results of (a) cumulative NPV, (b) cumulative CO$_2$ injection, and (c) cumulative brine production under the optimal well controls provided by DE, GPEME, and MSDRL.}
    \label{rsm}
\end{figure}

From the perspective of project management, we further analyzed the quality of the optimal well control schemes provided by DE, GPEME, and MSDRL. Later in this article, MSDRL refers to MSDRL (autotuned $\alpha$) unless otherwise stated. Fig.~\ref{rsm} shows the trend of the three important parameters over time, including cumulative NPV, cumulative CO$_2$ injection, and cumulative brine production. Compared to the scheme obtained by DE, the MSDRL scheme led to an 8.99\% increase in cumulative CO$_2$ injection and a 0.85\% increase in cumulative brine production. These shifts increased the final NPV by 13.76\%, from 1.526×10$^8$ USD to 1.736×10$^8$ USD. Compared to the scheme provided by GPEME, the MSDRL scheme resulted in a 5.14\% increase in cumulative CO$_2$ injection and a 3.86\% increase in cumulative brine production. This increased the final NPV by 5.82\%.

\subsection{Case 2: generalizable optimization scenario}

Having demonstrated the effectiveness and efficiency of MSDRL in deterministic optimization scenarios, we further applied it to tackle the generalizable optimization problem. In this situation, the agent trained on a large number of diverse scenarios is expected to leverage acquired knowledge to provide improved decisions for new scenarios quickly. This case is used to validate the generalization performance of MSDRL. The detailed network architectures for each module used in MSDRL can be found in Table~\ref{tableS3}-\ref{tableS7} in ~\ref{apc}.

\subsubsection{Evaluation of MLD model performance}
In this case, we constructed the MLD model using the results for 2000 simulation runs, each with different permeability fields, porosity fields, relative permeability curves, and well controls. These parameters were generated by sampling from different distributions over the allowable range. Detailed information on data generation and preprocessing can be found in ~\ref{apd}. We use the first 1600 samples as training data and the remaining 400 samples as test cases.

\begin{figure}[!htb]
    \centering
    \includegraphics[width=\textwidth]{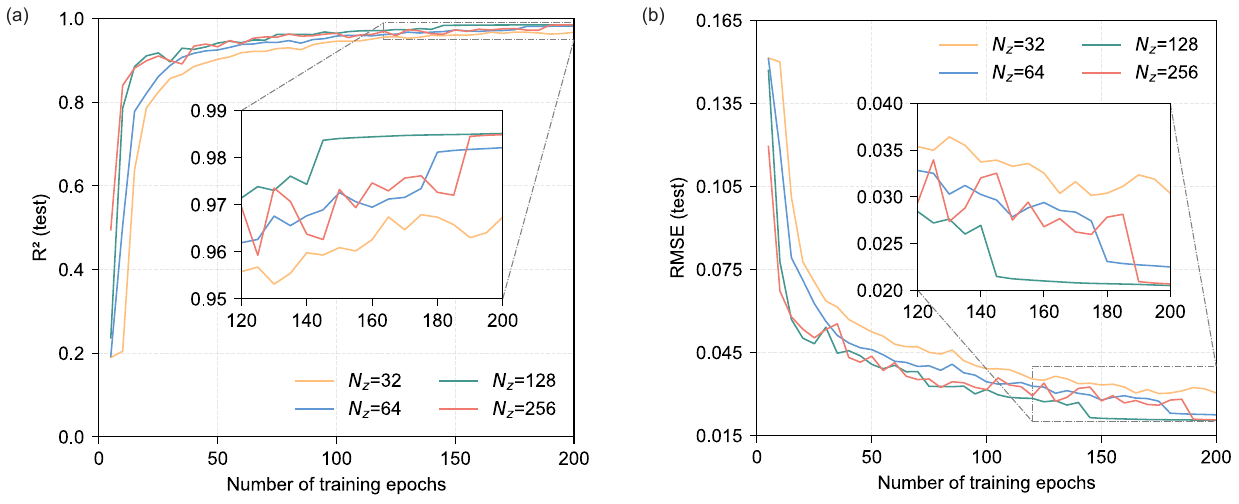}
    \caption{Evolution of R$^2$ and RMSE of models with different $N_z$ values evaluated on 400 test sample.}
    \label{assess2}
\end{figure}

In the context of generalizable optimization, the input of the MLD model involves different types of data. The MLD model learns a latent representation of them and evolves it in the latent space. Here the latent state dimension $N_z$ is crucial for the prediction performance of the model. We performed hyperparameter search over the latent state dimension of $\left\{ 32, 64, 128, 256 \right\}$ and used the model with the best test performance. Fig.~\ref{assess2} shows the evolution of R$^2$ and RMSE of models with different $N_z$ values evaluated on 400 test samples. It can be seen that the model with $N_z$=128 performs the best while exhibiting a more stable learning process. Thus, the model with $N_z$=128 was employed for forward flow predictions.

\begin{figure}[!htb]
    \centering
    \includegraphics[width=\textwidth]{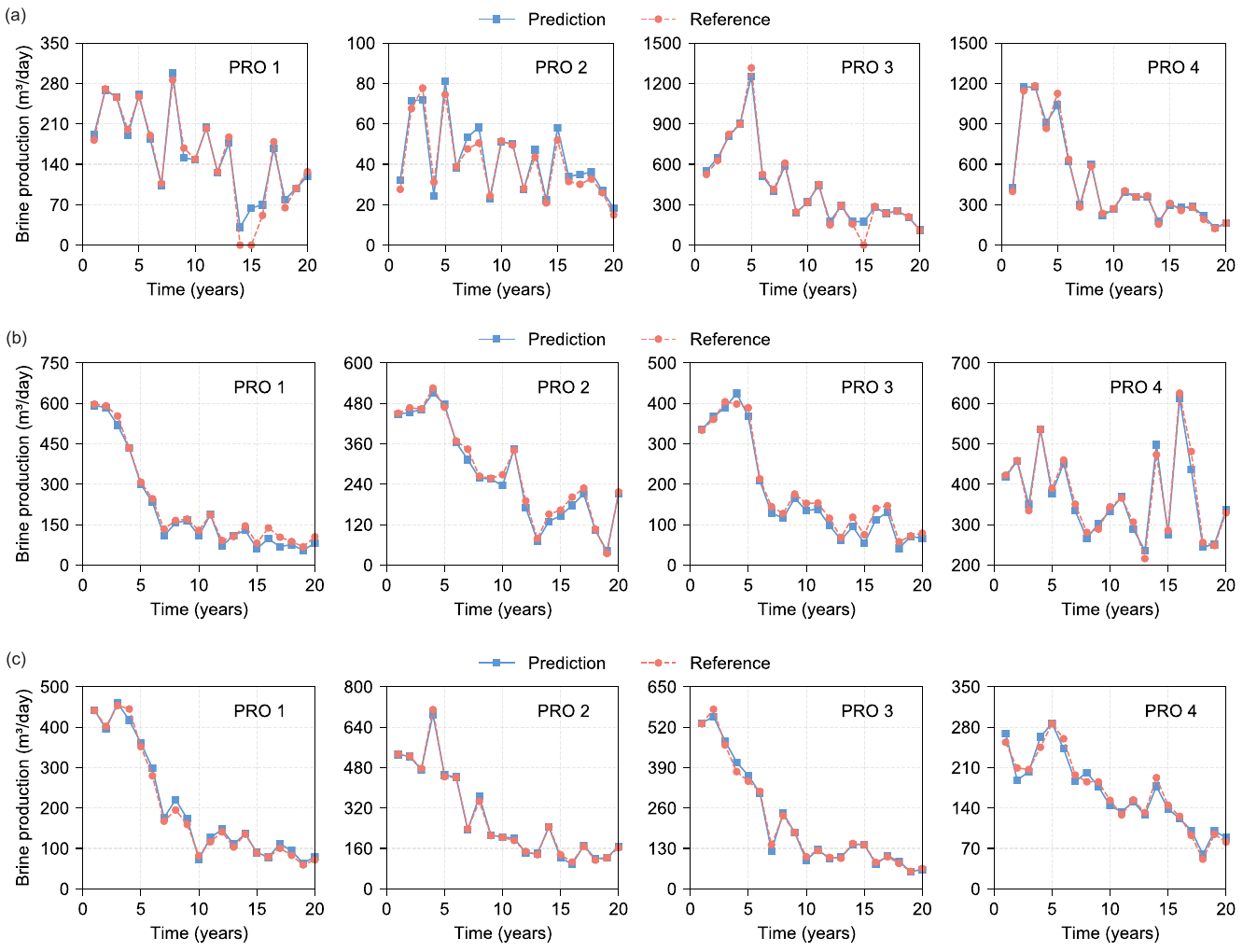}
    \caption{Comparison of brine production rates from the MLD model (blue lines) and the simulator (red lines) for (a) P$_{10}$, (b) P$_{50}$, and (c) P$_{90}$ R$^2$ test cases.}
    \label{brine}
\end{figure}

\begin{figure}[!htb]
    \centering
    \includegraphics[width=0.5\textwidth]{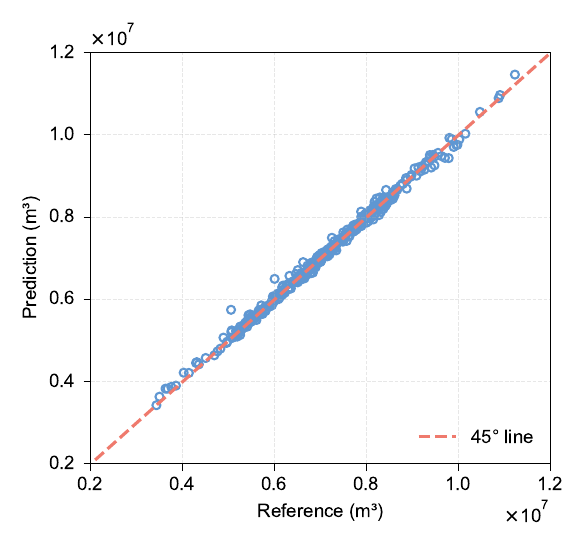}
    \caption{Test-case cross-plots for cumulative brine production.}
    \label{field}
\end{figure}

The R$^2$ score for each of the 400 test cases was calculated using Eq.~\ref{r2v}. The flow response predictions for P$_{10}$, P$_{50}$, and P$_{90}$ cases are displayed in Fig.~\ref{brine}. The blue line denotes the predictions from the MLD model and the red line denotes the references from the simulator. It can be seen that the prediction results can match the reference solutions well. Besides, since the MLD model can output the brine production rate for each producer at each time step, the cumulative brine production can be easily calculated. Fig.~\ref{field} displays the comparison of MLD prediction results and simulation results for all 400 test cases. It can be observed that most scatters fall near the 45° lines (red dotted line), which again illustrates the superior prediction performance of the learned MLD model for different scenarios.

\subsubsection{Optimization results analysis}

\begin{figure}[htb]\centering
    \includegraphics[width=0.95\textwidth]{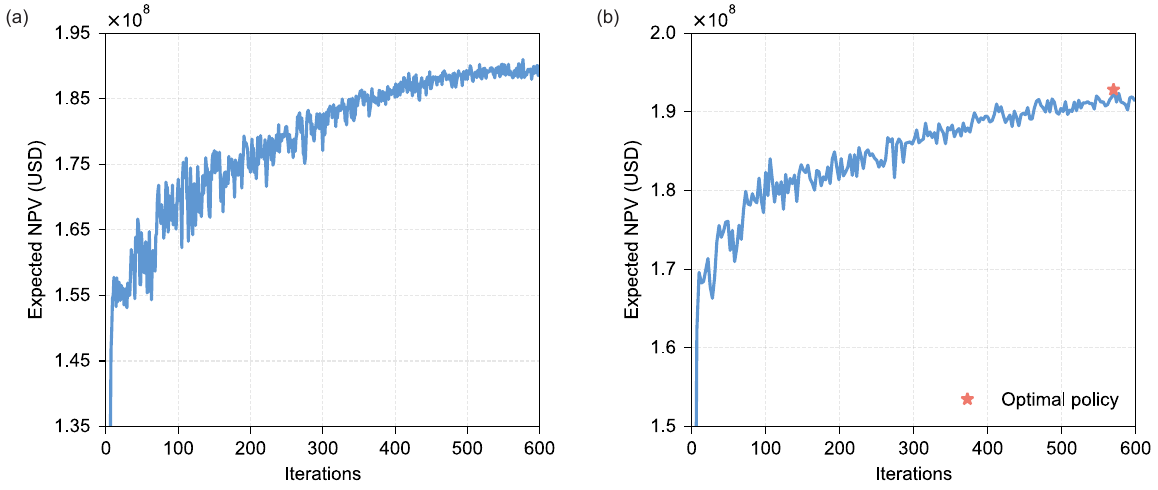}
    \caption{Policy optimization results. (a) Evolution of expected NPV in each training iteration. (b) Evolution of expected NPV for the test cases.}
    \label{opt2}
\end{figure}

The well control policy of the SAC agent here was trained with 1600 training scenarios, each with different permeability fields, porosity fields, and relative permeability curves. Using the learned MLD model as the environment, multiple interactions can be naturally parallelized in a single iteration through batch processing. This is in contrast to complex procedures that call the simulator in a distributed fashion. To alleviate the effects of randomness in training scenario generation and facilitate the monitoring of policy training, the expected NPV over 80 randomly selected scenarios was recorded after each iteration. The training process was terminated after 600 iterations. The evolution of the expected NPV during the training process is depicted in Fig.~\ref{opt2}a. It can be seen that the performance of the policy gradually improves as the training progresses. 

\begin{figure}[!htb]
    \centering
    \includegraphics[width=\textwidth]{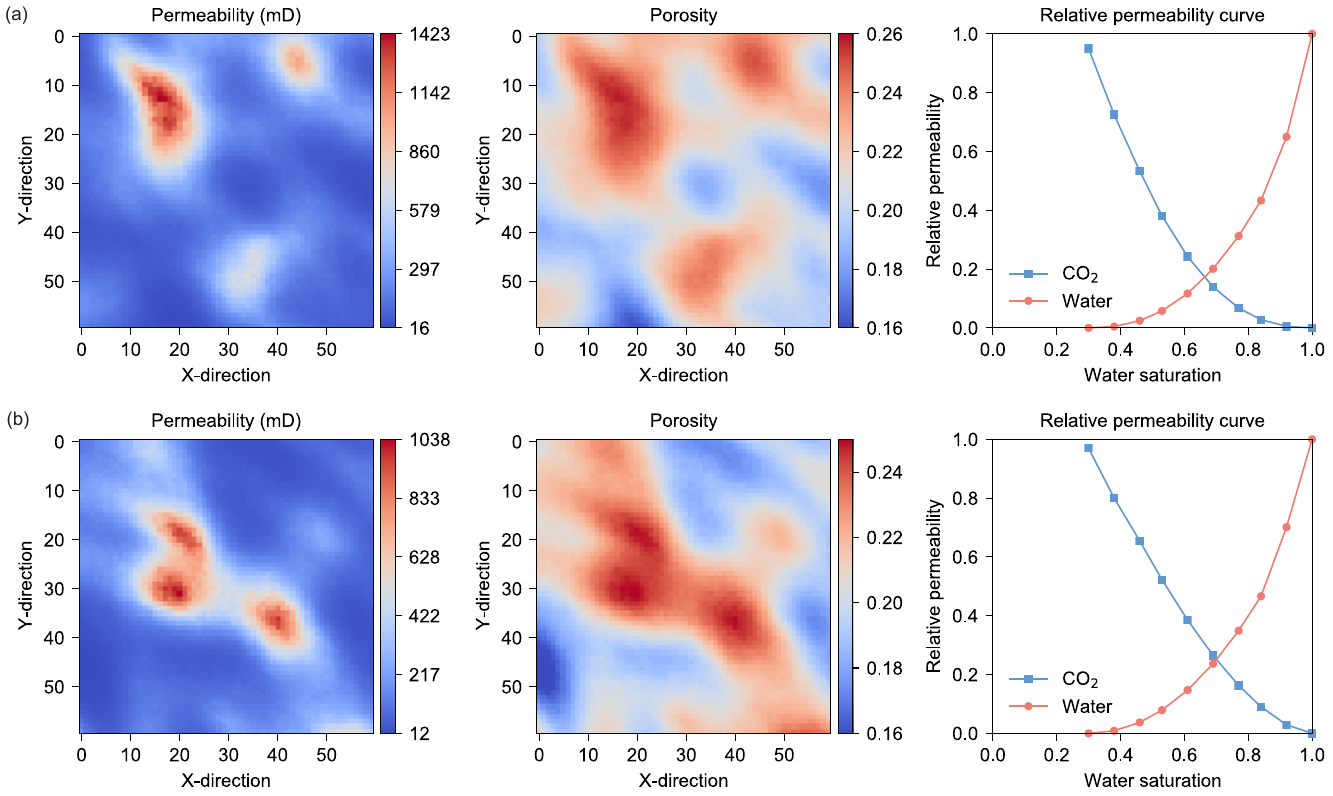}
    \caption{Permeability fields, porosity fields, and relative permeability curves for two randomly selected scenarios.}
    \label{sce}
\end{figure}

\begin{figure}[!htb]
    \centering
    \includegraphics[width=0.6\textwidth]{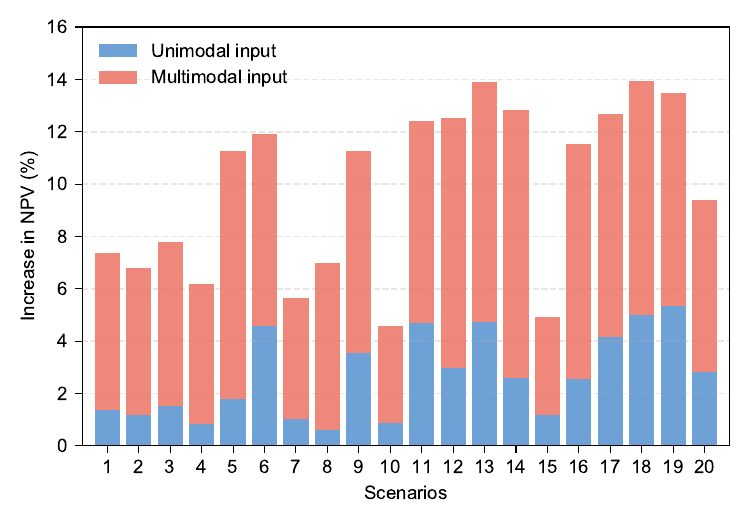}
    \caption{Benchmarking policy performance with random solutions.}
    \label{hist}
\end{figure}

We then assessed the performance of the trained policies with 20 new test scenarios, from which the model information of two randomly selected test scenarios is shown in Fig.~\ref{sce}. The evolution of expected NPV over these test scenarios, obtained by using the well controls from the most recent policy after every three iterations, is shown in Fig.~\ref{opt2}b. It can be seen that the test curve exhibits similar results to the training performance in Fig.~\ref{opt2}a. The optimal policy, shown as the red star, can be obtained after around 570 iterations. To further demonstrate how the multimodal inputs contribute to the generalization performance, we also tested the performance of the policy trained using unimodal inputs, which does not take into account the effect of permeability fields, porosity fields, and relative permeability curves during the optimization process. The performance of the control policies trained using these two modes was benchmarked with random (unoptimized) schemes. The evaluation results on 20 scenarios are shown in Fig.~\ref{hist}. It can be seen that the policy trained with multimodal input outperforms that trained with unimodal input to varying degrees. Compared to random schemes, they achieved an average improvement of 9.87\% and 2.66\%, respectively. Multi-modal inputs provide richer information about the environment, further improving the generalization performance of the policy.

\section{Discussion and conclusions}
\label{sec5}
In this work, we proposed a data-driven algorithm with superior efficiency and generalization ability, called MSDRL, for optimizing GCS operations. The major technical contributions include the design of the MLD model architecture and training strategy, as well as the application of advanced DRL to treat different types of well control optimization problems. The MLD model learns compressed latent representations of multimodal input features (i.e., dynamic state variables, static geologic parameters, fluid properties), and evolves the system states in the low-dimensional latent space while predicting the flow responses under time-varying well controls. To make it less susceptible to error accumulation, a training strategy combining a regression loss with a joint-embedding consistency loss was introduced to jointly optimize the above three modules. After offline training with available simulation data, the MLD model was integrated with the state-of-the-art DRL algorithm SAC to automatically discover the optimal control policies. MSDRL provides a new insight for unifying multimodal deep learning models and DRL into a single framework, by learning a compressed latent representation and then training DRL agents in the learned latent space.

The proposed MSDRL algorithm was first applied to a 3D deterministic GCS optimization case and its performance was tested against commonly used evolutionary algorithm DE and surrogate-assisted algorithm GPEME. It was shown that MSDRL achieved the best results in terms of both effectiveness and efficiency. Specifically, compared to the schemes obtained by DE and GPEME, the MSDRL scheme leads to a 13.73\% and 5.82\% increase in final NPV, respectively, while reducing the required computational cost by more than 60\% (30 hours). The framework was further employed for a generalizable optimization case, showing that the agent exhibits excellent generalizable performance and can utilize the gained knowledge to quickly provide improved decisions for new scenarios.

Considering the above promising results, we would like to pursue further extensions to this work. First, the training and test scenarios used in this study were generated from simple distributions, whereas real-world scenarios are more complex. For practical applications, more diverse scenarios and richer geological parameters need to be involved in the training process. Besides, the present approach builds the MLD model in an offline manner, that is, the model was trained using a static dataset. For more complicated problems, an online iterative training procedure may be required. This involves dynamically updating the model parameters using newly collected data so that the model can be improved and refined over time. It is also of interest to integrate advanced constraint processing techniques into the framework. In this situation, the agent not only seeks to maximize the NPV rewards but also satisfies the nonlinear state constraints. Finally, MSDRL provides a general framework for incorporating other DRL algorithms. For example, we could replace SAC with deep Q-network \citep{mnih2015human} and then apply it to solve discrete optimization problems as demonstrated in \citep{sun2020optimal}.

\section*{Acknowledgment}

\noindent{This work is supported by the High Performance Computing Center at Eastern Institute of Technology, Ningbo, the National Center for Applied Mathematics Shenzhen (NCAMS), the Shenzhen Key Laboratory of Natural Gas Hydrates (Grant No. ZDSYS20200421111201738), the SUSTech – Qingdao New Energy Technology Research Institute, the National Natural Science Foundation of China (Grant No. 62106116), and Natural Science Foundation of Ningbo of China (No. 2023J027).}


\appendix
\section{Training details of SAC algorithm}
\label{apa}
\setcounter{equation}{0}
\renewcommand\theequation{A.\arabic{equation}}

SAC is based on the policy iteration, where the policy evaluation step and the policy improvement step alternate until the iteration converges to the optimal policy.

In the policy evaluation step, the parameters ${{\mathsf{\psi }}_{j}}$ of the Q-function can be trained by minimizing the loss function ${{\mathcal L}_{Q}}({{\mathsf{\psi }}_{j}})$ (for $j=1,2$):
\begin{equation}
    {{\mathcal L}_{Q}}({{\mathsf{\psi }}_{j}})={{\mathbb{E}}_{\left( {{\mathbf{s}}_{t}},{{\mathbf{a}}_{t}},{{r}_{t}},{{\mathbf{s}}_{t+1}} \right)\sim \mathcal{D}}}\left[ {{\left( {{Q}_{{{\mathsf{\psi }}_{j}}}}\left( {{\mathbf{s}}_{t}},{{\mathbf{a}}_{t}} \right)-y\left( {{r}_{t}},{{\mathbf{s}}_{t+1}} \right) \right)}^{\text{2}}} \right]
    \label{seq20}
\end{equation}

The parameters ${{\mathsf{\psi }}_{\text{targ},j}}$ of target Q-networks are updated according to polyak averaging:
\begin{equation}
    {{\mathsf{\psi }}_{\text{targ},j}}\leftarrow \tau {{\mathsf{\psi }}_{j}}+\left( 1-\tau  \right){{\mathsf{\psi }}_{\text{targ},j}}
    \label{seq21}
\end{equation}
where $\tau $ is the smoothing factor.

In the policy improvement step, the parameters $\mathsf{\varphi }$ of the policy can be trained by minimizing the loss function ${{\mathcal L}_{\pi }}(\mathsf{\varphi })$:
\begin{equation}
    {{\mathcal L}_{\pi }}(\mathsf{\varphi })={{\mathbb{E}}_{{{\mathbf{s}}_{t}}\sim \mathcal{D},{{\mathbf{a}}_{t}}\sim {{\pi }_{\mathsf{\varphi }}}\left( \cdot |{{\mathbf{s}}_{t}} \right)}}\left[ \alpha \log \left( {{\pi }_{\mathsf{\varphi }}}\left( {{\mathbf{a}}_{t}}|{{\mathbf{s}}_{t}} \right) \right)-\underset{j=1,2}{\mathop{\min }}\,{{Q}_{{{\mathsf{\psi }}_{j}}}}\left( {{\mathbf{s}}_{t}},{{\mathbf{a}}_{t}} \right) \right]
    \label{seq22}
\end{equation}

During training, the temperature parameter $\alpha $ can be automatically learned with:
\begin{equation}
    \mathcal L(\alpha )={{\mathbb{E}}_{{{\mathbf{a}}_{t}}\sim {{\pi }_{\mathsf{\varphi }}}}}\left[ -\alpha \log {{\pi }_{\mathsf{\varphi }}}\left( {{\mathbf{a}}_{t}}\mid {{\mathbf{s}}_{t}} \right)-\alpha {{\mathcal{H}}_{0}} \right]
    \label{seq23}
\end{equation}
where ${{\mathcal{H}}_{0}}$ is a constant, which is usually set to the dimension of the action space.

\section{Hyperparameter configurations for MSDRL}
\label{apb}
\renewcommand\thetable{B.\arabic{table}} 
\setcounter{table}{0}

We list the most important hyperparameters for the MLD model and SAC algorithm.

\begin{table}[!htb]
\small
\centering
\caption{Hyperparameter for MLD.}
\label{tableS1}
\begin{tabular}{ccc}
\hline
Hyperparameter                 & Value \\ \hline
latent state dimension & 128   \\
Training epoch            & 200   \\
Batch size                & 25    \\
Learning rate             & 0.001  \\
Optimizer                 & Adam   \\
Weight decay              & 0.0005  \\ \hline
\end{tabular}
\end{table}

\begin{table}[!htb]
\small
\centering
\caption{Hyperparameter for SAC.}
\label{tableS2}
\begin{tabular}{ccc}
\hline
Hyperparameter                 & Value \\ \hline
Discount factor & 0.95   \\
Smoothing factor            & 0.005   \\
Batch size                & 128    \\
Replay buffer size        & 20000    \\
Learning rate             & 0.0003  \\
Optimizer                 & Adam  \\  \hline
\end{tabular}
\end{table}

\section{Neural network architecture details}
\label{apc}
\renewcommand\thetable{C.\arabic{table}} 
\setcounter{table}{0}
MSDRL consists of a total of five submodules: representation module, transition module, prediction module, actor, and critic. The detailed architectures of these modules are
provided in Table \ref{tableS3}-\ref{tableS7}.

\begin{table}[H]
\small
\centering
\caption{The network architecture of the representation module. Conv and FC denote the convolution and fully connected operation, respectively. ReLU and Tanh denote the activation function.}
\label{tableS3}
\begin{tabular}{ccc}
\hline
Layer          & Configuration                                           & Output Shape   \\ \hline
               & \textbf{Branch 1: encoding permeability   and porosity} &                \\
Input 1        & ---                                                     & (N,2×3,60,60)  \\
Conv 1-1       & 32 filters of size 3×3×3, stride=2, activation=ReLU      & (N,32,29,29)   \\
Conv 1-2       & 32 filters of size 3×3×3, stride=2, activation=ReLU      & (N,32,14,14)   \\
Conv 1-3       & 64 filters of size 3×3×3, stride=2, activation=ReLU      & (N,64,6,6)     \\
Conv 1-4       & 64 filters of size 3×3×3, stride=2, activation=ReLU      & (N,64,2,2)     \\
Flatten        & ---                                                     & (N,256)        \\
Output 1       & Neurons=256, activation=ReLU                            & (N,256)        \\
               & \textbf{Branch 2: encoding pressure   and saturation}   &                \\
Input 2        & ---                                                     & (N,2×3,60,60)  \\
Conv 2-1       & 32 filters of size 3×3×3, stride=2, activation=ReLU      & (N,32,29,29)   \\
Conv 2-2       & 32 filters of size 3×3×3, stride=2, activation=ReLU      & (N,32,14,14)   \\
Conv 2-3       & 64 filters of size 3×3×3, stride=2, activation=ReLU      & (N,64,6,6)     \\
Conv 2-4       & 64 filters of size 3×3×3, stride=2, activation=ReLU      & (N,64,2,2)     \\
Flatten        & ---                                                     & (N,256)        \\
Output 2       & Neurons=256, activation=ReLU                            & (N,256)        \\
               & \textbf{Branch 3: encoding relative   permeability}     &                \\
Input 3        & ---                                                     & (N,6)          \\
FC             & Neurons=64, activation=ReLU                             & (N,64)         \\
Output 3       & Neurons=64, activation=ReLU                             & (N,64)         \\
               & \textbf{Concat}                                         &                \\
Output 1, 2, 3 & ---                                                     & (N,256+256+64) \\
Output         & Neurons=128, activation=Tanh                            & (N,128)        \\ \hline
\end{tabular}
\end{table}

\begin{table}[H]
\small
\centering
\caption{The network architecture of the transition module.}
\label{tableS4}
\begin{tabular}{ccc}
\hline
Layer  & Configuration                & Output Shape \\ \hline
Input  & ---                          & (N,128+5)    \\
FC     & Neurons=256, activation=ReLU & (N,256)      \\
Output & Neurons=128, activation=Tanh & (N,128)      \\ \hline
\end{tabular}
\end{table}

\begin{table}[H]
\small
\centering
\caption{The network architecture of the prediction module.}
\label{tableS5}
\begin{tabular}{ccc}
\hline
Layer  & Configuration                & Output Shape \\ \hline
Input  & ---                          & (N,128+5)    \\
FC     & Neurons=256, activation=ReLU & (N,256)      \\
Output & Neurons=4, activation=Tanh   & (N,4)        \\ \hline
\end{tabular}
\end{table}

\begin{table}[H]
\small
\centering
\caption{The Network architecture of the actor.}
\label{tableS6}
\begin{tabular}{ccc}
\hline
Layer                  & Configuration                                  & Output Shape \\ \hline
Input                  & ---                                            & (N,128)      \\
FC 1                   & Neurons=256, activation=ReLU                   & (N,256)      \\
FC 2                   & Neurons=256, activation= ReLU                  & (N,256)      \\
FC 3 ($\mu $)          & Neurons=5, activation=ReLU                     & (N,5)        \\
FC 4 ($\ln (\sigma )$) & Neurons=5, activation= ReLU                    & (N,5)        \\
Output                 & Action=$Normal(\mu ,\sigma )$, activation=Tanh & (N,5)        \\ \hline
\end{tabular}
\end{table}

\begin{table}[H]
\small
\centering
\caption{The network architecture of the critic.}
\label{tableS7}
\begin{tabular}{ccc}
\hline
Layer  & Configuration                 & Output Shape \\ \hline
Input  & ---                           & (N,128+5)    \\
FC 1   & Neurons=256, activation=ReLU  & (N,256)      \\
FC 2   & Neurons=256, activation= ReLU & (N,256)      \\
Output & Neuron=1                      & (N,1)        \\ \hline
\end{tabular}
\end{table}

\section{Data preparation}
\label{apd}
\setcounter{equation}{0}
\renewcommand\theequation{D.\arabic{equation}}

\noindent{\textbf{(1) Data Generation.} We refer to previous studies to generate different types of parameters.}
\begin{itemize}[leftmargin=*]
\item Permeability $\mathbf{k}$. The permeability field (the unit is mdarcy, mD) is assumed to follow a log-normal distribution. The mean value and standard deviation value are set to 5.0 and 1.0, respectively. We use Stanford Geostatistical Modeling Software \citep{remy2009applied} to generate required permeability realizations.
\item Porosity $\mathsf{\phi }$. The porosity is simply assumed to be correlated with permeability, which can be calculated using the following equation \citep{zhong2019predicting}.
\begin{equation} 
    \mathsf{\phi }=\text{0}\text{.05lo}{{\text{g}}_{\text{10}}}(\mathbf{k})+\text{0}\text{.1}
    \label{eqb1}
\end{equation}
\item Relative permeability ${{k}_{r}}$. The relative permeability is a nonlinear function of saturation, which can be described using the modified Brooks-Corey (MBC) model \cite{goda2004using}. From this model, the expression of ${{k}_{r}}$ can be determined with vector $\left[ k_{rg}^{0},k_{rw}^{0},{{S}_{gr}},{{S}_{wc}},{{n}_{g}},{{n}_{w}} \right]$ \citep{zhang2022predictionsurro}. We use the Latin hypercube sampling to generate the required samples within reasonable ranges.
\item Well control $\mathbf{a}$. We use the Latin hypercube sampling to generate 2000 groups of well controls. Each includes the schemes of all producers and injectors at 20 control steps.
\end{itemize}

High-fidelity simulations are performed for different combinations of parameters. The dynamic state variables (e.g., pressure fields and saturation fields) and flow responses at 20 control steps are saved after simulation. Note that this process is easy to implement in parallel if multiple CPU cores are available. Through the above process, a raw dataset containing 2000 samples is constructed. 
\vspace{0.5em}

\noindent{\textbf{(2) Data normalization.} 
\vspace{0.5em}
Data normalization is an effective way to improve the robustness of DNNs. Before training, we use Eq.~\ref{eqb4} to normalize the data.
\begin{equation} 
    f\left( x \right)=\text{2}\times \left[ \frac{x-\min \left( x \right)}{\max \left( x \right)-\min \left( x \right)} \right]-\text{1}
    \label{eqb4}
\end{equation}

It can be observed that after normalization, all data is restricted to the range of -1 to 1.

\end{document}